\documentclass[journal]{IEEEtran}
\usepackage{cite}
\ifCLASSINFOpdf
\else
\fi
\usepackage{comment}
\usepackage{hyperref}
\usepackage{verbatim}
\usepackage{amsmath,graphicx}

\usepackage{mathrsfs}
\usepackage{amsopn}
\usepackage{cite}
\usepackage{amsthm}
\usepackage{epsfig,psfrag}
\usepackage{amsfonts,amssymb}
\usepackage{subcaption}
\usepackage{epsfig}
\usepackage{epstopdf}
\usepackage{amsbsy}
\usepackage{tabularx}
\usepackage{amssymb} 
\usepackage{multirow}
\usepackage{xcolor}
\usepackage{array}
\usepackage{arydshln}
\usepackage{color}
\usepackage{caption}
\usepackage{xcolor}
\usepackage{nicefrac}
\usepackage{cleveref}
\usepackage{graphics}
\usepackage{algorithm}
\usepackage{algorithmic}
\usepackage{amsmath}
\usepackage{xcolor}
\usepackage{xfrac}
\usepackage{multirow}
\usepackage{booktabs}
\graphicspath{{figures/}}
\newcommand{\bb}{\boldsymbol }

\newcommand{\NO}{\textcolor[rgb]{0,0,0}}
\newcommand{\mgt}{\textcolor[rgb]{0,0,0}}
\newcommand{\hl}{\textcolor[rgb]{0,0,0}}
\newcommand{\ff}{\textcolor[rgb]{0,0,0}}
\begin{document}

\title{Coupled Feature Learning for Multimodal Medical Image Fusion}
\author{Farshad G. Veshki,~\IEEEmembership{Student Member,~IEEE,} Nora Ouzir,~\IEEEmembership{Member,~IEEE,} Sergiy A. Vorobyov,~\IEEEmembership{Fellow,~IEEE,} and Esa Ollila,~\IEEEmembership{Senior Member,~IEEE}
\thanks{Farshad~G.~Veshki, S.~A.~Vorobyov and Esa~Ollila are with the Department of Signal Processing and Acoustics, Aalto University, Espoo, Finland. Nora Ouzir is with Department of Mathematics, CentraleSupelec, France (e-mail: farshad.ghorbaniveshki@aalto.fi; nora.ouzir@centraleSupelec.fr; sergiy.vorobyov@aalto.fi; esa.ollila@aalto.fi). \textit{Corresponding author: S.~A.~Vorobyov.}}}

\maketitle
\begin{abstract}
Multimodal image fusion aims to combine relevant information from images acquired with different sensors. In medical imaging, fused images play an essential role in both standard and automated diagnosis. In this paper, we propose a novel multimodal image fusion method based on coupled dictionary learning. The proposed method is general and can be employed for different medical imaging modalities. Unlike many current medical fusion methods, the proposed approach does not suffer from intensity attenuation nor loss of critical information. Specifically, the images to be fused are decomposed into coupled and independent components estimated using sparse representations with identical supports and a Pearson correlation constraint, respectively. An alternating minimization algorithm is designed to solve the resulting optimization problem. The final fusion step uses the max-absolute-value rule. Experiments are conducted using various pairs of multimodal inputs, including real MR-CT and MR-PET images. The resulting performance and execution times show the competitiveness of the proposed method in comparison with state-of-the-art medical image fusion methods.
\end{abstract} 

\IEEEpeerreviewmaketitle

\section{Introduction}\label{section_introduction}

\IEEEPARstart{M}{edical} imaging techniques are designed to capture specific types of tissues and structures inside the body. Anatomical imaging techniques can provide high-resolution images of internal organs. For example, magnetic resonance (MR) imaging shows soft tissues, such as fat and liquid, while computed tomography (CT) captures more effectively hard tissues, such as bones and implants. Functional imaging techniques measure the biological activity of specific regions inside the organs. Single-photon emission computed tomography (SPECT) and positron emission tomography (PET) are typical examples of this type of techniques. The goal of multimodal image fusion is to combine this variety of (often complementary) information into a single image. In addition to easing the visualisation of multiple images, fusion enables a joint analysis providing relevant and new information about the patient, for both traditional and automated diagnosis~\cite{survey2004,survey2017,overview2016}.

\NO{Medical image fusion methods mostly rely on the extraction of different types of features in the images before selecting or combining the most relevant ones. One way of achieving this is by transforming the images into a domain where the relevant features would arise naturally. A common approach employs multi-scale transformation (MST) techniques to extract features from different levels of resolution and select them using an appropriate fusion rule in a subsequent step~\cite{NSST-PAPCNN2019, ULAP2016, NSCT-PCDC2013, LP-SR2015, LLF-IOI2017, PSF2018,Jiao2016}. The final fused image is obtained by applying the inverse MST to the combined multi-scale features. For example, a recent MST-based fusion method using a non-subsampled shearlet transform (NSST) has been proposed in~\cite{NSST-PAPCNN2019}. The fusion rule is based on a pulse coupled neural network (PCNN), weighted local energy and a modified Laplacian~\cite{NSST-PAPCNN2019}. In a different study, the MST is based on local Laplacian filtering (LLF), and the fusion rule relies on the information of interest (IOI) criterion\cite{LLF-IOI2017}.}

\NO{Another approach seeks to learn the relevant features from the images. The methods that follow this approach mainly use sparse representations (SR) and dictionary learning~\cite{CSR2016,CSMCA2019,SRMCA2014,SRfusion2018,SR-SOMP2012,SR_fusion2010,SR_fusion2011}. First, dictionaries are learnt and used to find an SR of the images. The resulting \textit{sparse codes} from both input images are then selected by measuring their respective \textit{levels of activity} (\textit{e.g.,} using the $\ell_1-$norm). Once the combined SR is obtained, the fused image is reconstructed again using the learnt dictionaries. In this category of methods, different strategies \NO{have been} proposed to deal with the presence of drastically different components in the images. For example, the images are separated into base and detail components prior to the dictionary learning phase in~\cite{CSR2016}. In \cite{CSMCA2019,SRMCA2014}, morphological characteristics, such as cartoon and texture components, are used.}

\NO{The multi-component strategy is useful for both SR and MST-based fusion. It allows the different layers of information to be identified before extracting the relevant (multi-level) features, \mgt{and thus reduces the loss of contrast or image quality}. Many works have shown the superiority of this approach compared to using a single layer or component in each image~\cite{NSST-PAPCNN2019,LLF-IOI2017,PSF2018,SRfusion2018}. However, the choice of the multi-component model is not straightforward because all decompositions do not necessarily yield features that are suitable for fusion. For example, in the MST approach, the highest resolution level of MR-CT images usually depicts tissues with very different characteristics, \NO{\textit{i.e.,} soft tissues in MR and hard structures in CT}. Fusing \NO{soft and hard tissues} (using binary selection, for example) can \NO{cause} a loss of useful information. \mgt{Using averaging techniques can mitigate this loss, but it leads to an attenuation of the original intensities (particularly when there is a weak signal in one of the input images).} More generally, it is not always meaningful to apply a fusion rule when there is no guarantee features with the same levels of resolution encompass the same type of information. Conversely, different imaging techniques can provide varying resolutions for the same underlying structures. Finally, it is not always clear in practice how to choose the number of resolution levels for MST or an adequate basis for the SR-approach.} 

\NO{This paper presents a novel multi-modal image fusion method using an SR-based multi-component approach. A general decomposition model enables us to preserve the important features in both input modalities and reduce the loss of essential information. Specifically, coupled dictionary learning (CDL) is used to learn features from input images and simultaneously decompose them into correlated and independent components. The core idea of the proposed model is that the independent components contain modality-specific information that should appear in the final fused image. Therefore, these components should be preserved entirely rather than subjected to some fusion rule. Since the input images still represent the same anatomical region or organ, they can contain a significant amount of similar or overlapping information.} \NO{This information is taken into account in the correlated components and considered relevant for fusion.} 
\NO{The coupled dictionaries play a key role here; each couple of atoms represent a correlated feature. This allows us to choose the best candidate for fusion based on the most significant dictionary atom without any loss of information. A summary of the proposed methodology and contributions is provided in the following}
\begin{itemize}
    \item We employ a general decomposition model suitable for fusing images from various medical imaging modalities.
    \item A CDL method based on simultaneous sparse approximation is proposed for estimating the correlated features. In order to incorporate variability in the appearance of correlated features, we relax the assumption of equal SRs by assuming common supports only. 
    \item The independent components are estimated using a Pearson correlation-based constraint (enforcing low correlation).
    \item An alternating optimization method is designed for simultaneous dictionary learning and image decomposition.
    \item \NO{The final fusion step combines direct summation with the max-absolute-value rule.}
\end{itemize}

\NO{A thorough experimental comparison with current medical image fusion methods is conducted using multiple pairs of real multimodal images. The data comprises six different combinations of medical imaging techniques, including MR-CT, MR-PET and MR-SPECT. The experimental results show that the proposed method results in a better fusion of local intensity and texture information, compared to other current techniques. In particular, isolating modality-specific information reduces the loss of information significantly.}

\subsubsection*{Notations} Throughout the paper, we use bold capital letters for matrices. A matrix with a single subscript always denotes a column of that matrix. For example $[\boldsymbol{D}_1]_i$ is the $i-$th column of $\boldsymbol{D}_1$. In a matrix, the entry at the intersection of $i$th row and $j$th column is denoted as $[\cdot]_{(i,j)}$. In addition, the Frobenius norm of a matrix is denoted by $\|\cdot\|_{F}$, the Euclidean norm of a vector is denoted by $\|\cdot\|_2$, and $\|\cdot\|_0$ is the operator counting the number of nonzero coefficients of a vector. The $\mathrm{supp}\{\cdot\}$ denotes the support of a matrix. Operator $|\cdot|$ denotes the absolute value of a number. The symbol $(\cdot)^T$ denotes the transpose operation and $(\cdot)^+$ stands for the updated variable. The symbol $\perp \!\!\! \perp$ denotes the conditional independence between two variables.

The remainder of the paper is organised as follows. Section~\ref{section_CDL} presents the proposed CDL approach and SR with common supports. The proposed image decomposition method and the fusion step are explained in Section~\ref{section_Decomposition} and \ref{section_fusion}, respectively. Section~\ref{section_results} reports the experimental results using various examples multi-modal images. Finally, conclusions are provided in Section~\ref{section_conclusion}.

\section{Coupled Feature Learning}
\label{section_CDL}
The objective of CDL is to learn a pair of dictionaries $\boldsymbol{D}_1$ and $\boldsymbol{D}_2$, used to jointly represent two datasets $\boldsymbol{X}_1$ and $\boldsymbol{X}_2$. \NO{The underlying relationship between these datasets is captured using} a common sparse coding matrix $\boldsymbol{A}$. A standard formulation of the CDL problem is given by the following minimization problem 
\begin{equation}
\begin{split}
\begin{aligned}
&\underset{ {\boldsymbol{D}_1,\boldsymbol{D}_2, \boldsymbol{A} } }{\mathrm{min}} \;\left \| \boldsymbol{D}_1 \boldsymbol{A} - \boldsymbol{X}_1 \right\| _{\rm F}^2 +\left\|\boldsymbol{D}_2 \boldsymbol{A}-\boldsymbol{X}_2 \right\| _{\rm F}^2 \\ 
&{\text{s.t.}} \left\| \boldsymbol{A}_i \right\| _0 \leq T,\; \left\| [\boldsymbol{D}_1]_t \right\|_{2}= 1,\; \left\| [\boldsymbol{D}_2]_t \right\|_{2}= 1,\;\forall t,i\\
\label{std_CDL}
\end{aligned} 
\end{split}
\end{equation} 
\noindent where $T$ is the maximum number of non-zero coefficients in each column of the sparse coding matrix $\boldsymbol{A}$. The constraint on the norm of the dictionary elements is used to avoid scaling ambiguities. \NO{CDL is particularly suitable for tackling problems that involve image reconstruction in different feature spaces. The standard problem \eqref{std_CDL} has been successfully employed in numerous image processing applications, such as image super-resolution~\cite{SR_CDL2012}, single-modal image fusion~\cite{CDL_fusion2020}, or photo and sketch mapping~\cite{Sketch_CDL2012}.} 
\NO{In this section, the objective is to learn} the correlated features in two input multi-modal images. These features are captured \NO{by the atoms} of the learnt dictionaries $\boldsymbol{D}_1$ and $\boldsymbol{D}_2$. \ff{Since we are dealing with images acquired from different sensors, different aspects of the same underlying structure can be displayed with different levels of visibility in each modality.} For example, both CT and MRI can show tendons, but they are more visible in MR images~\hl{\cite{MRI-tendon2018}. We incorporate this aspect by imposing identical supports instead of enforcing an equal sparse representation for each pair of dictionary atoms. In this way, correlated elements can be represented with different levels of significance in each dictionary. The proposed} modified CDL problem can be formulated as follows
\begin{equation}
\begin{split}
\begin{aligned}
&\underset{ {\boldsymbol{D}_1,\boldsymbol{D}_2, \boldsymbol{A}_1, \boldsymbol{A}_2} }{\mathrm{minimize}} \left \| \boldsymbol{D}_1 \boldsymbol{A}_1 - \boldsymbol{X}_1 \right\| _{\rm F}^2+\left\|\boldsymbol{D}_2 \boldsymbol{A}_2-\boldsymbol{X}_2 \right\| _{\rm F}^2 \\
&{\text{s.t.}} \quad \operatorname{supp}\{\boldsymbol{A}_1\} = \operatorname{supp}\{\boldsymbol{A}_2\}\\
& \qquad \left\| [\boldsymbol{A}_1]_i \right\| _0 \leq T, \left\| [\boldsymbol{A}_2]_i \right\| _0 \leq T, \forall i\\
& \qquad \left\| [\boldsymbol{D}_1]_t \right\|_{2}= 1, \left\| [\boldsymbol{D}_2]_t \right\|_{2}= 1, \forall t.
\end{aligned} \label{cor_estimation_CDL}
\end{split}
\end{equation} 
\noindent \NO{Problem \eqref{std_CDL} is typically solved by alternating between a sparse coding stage and a dictionary update step~\cite{KSVD2006,MOD1999,ODL2009}. In this work, we solve problem \eqref{cor_estimation_CDL} using an alternating approach too. } Specifically, after splitting the variables into two subsets $\{\boldsymbol{A}_1, \boldsymbol{A}_2\}$ and $\{\boldsymbol{D}_1, \boldsymbol{D}_2\}$, we alternate between two optimization phases detailed in the following subsections.   
\subsection{Sparse Coding}
Minimizing \eqref{cor_estimation_CDL} with respect to the first set of variables $\{\boldsymbol{A}_1, \boldsymbol{A}_2\}$ requires some changes to the standard sparse coding procedure \hl{(\textit{e.g.}, OMP~\cite{OMP1993})}. First, one has to enforce common supports. The method of simultaneous orthogonal matching pursuit (SOMP)~\cite{SOMP2005} is of interest here \NO{as it includes a common support constraint. Secondly, one has to consider coupled dictionaries.} This can be achieved by modifying the atom selection rule of SOMP so that each of the input signals is approximated using a different dictionary instead of sharing a single one. In each iteration, this modified algorithm selects a pair of coupled atoms that minimizes the sum of the residual \NO{errors}, \textit{i.e.}, 
\begin{equation*}
\underset{t}{\mathrm{argmax}} | \boldsymbol{r}_1^T [\boldsymbol{D}_1]_t | + |\boldsymbol{r}_2^T [\boldsymbol{D}_2]_t |
\end{equation*}
where $\boldsymbol{r}_1$ and $\boldsymbol{r}_2$ are the approximation residuals of a pair of input signals (\textit{e.g.}, $\boldsymbol{r}_1 = [\boldsymbol{X}_1]_i - \boldsymbol{D}_1[\boldsymbol{A}_1]_i$ and $\boldsymbol{r}_2 = [\boldsymbol{X}_2]_i - \boldsymbol{D}_2[\boldsymbol{A}_2]_i$ are the residuals corresponding to the $i$th columns of $\boldsymbol{X}_1$ and $\boldsymbol{X}_2$, respectively). \NO{The optimal nonzero coefficients of the sparse codes are then computed based on their associated selected atoms. As opposed to} SOMP, we stop the algorithm when only one of the input signals meets the stopping criterion \NO{(\textit{i.e.,}} the Euclidean norm of one of the residuals is smaller than a \NO{user-defined} threshold $\epsilon$).
\hl{This is based on the fact that the main objective of the algorithm is to estimate correlated features, whereas any remaining noise (once the approximation of one of the signals is complete) is evidently uncorrelated with the residual of the second signal.}
\subsection{Dictionary Update}
The first two terms of \eqref{cor_estimation_CDL} are independent with respect to the dictionaries $\boldsymbol{D}_1$ and $\boldsymbol{D}_2$. They are also independent with respect to the non-zero coefficients in $\boldsymbol{A}_1$ and $\boldsymbol{A}_2$ when the supports are fixed. Therefore, the dictionaries $\boldsymbol{D_1}$ and $\boldsymbol{D_2}$ can be updated individually using \NO{any efficient} dictionary learning algorithm. \NO{We choose to use the K-SVD} method~\cite{KSVD2006} because it updates \NO{the dictionary atoms and the associated sparse coefficients without changing their supports. Specifically, K-SVD uses a singular value decomposition.}

\begin{figure}[!h]
\centering
\begin{subfigure}{.19\textwidth}
  \centering
  \includegraphics[width=1\linewidth]{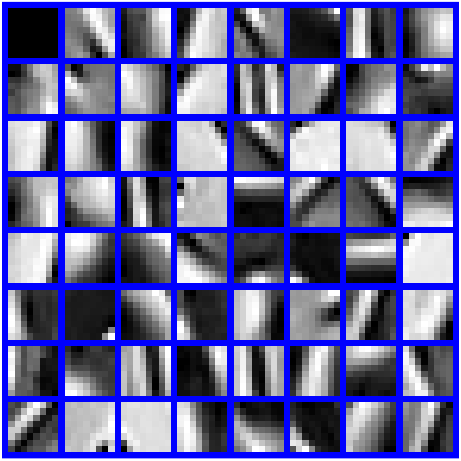} 
  \caption{}
  \label{fig: decomposition D1}
\end{subfigure}
\begin{subfigure}{.19\textwidth}
  \centering
  \includegraphics[width=1\linewidth]{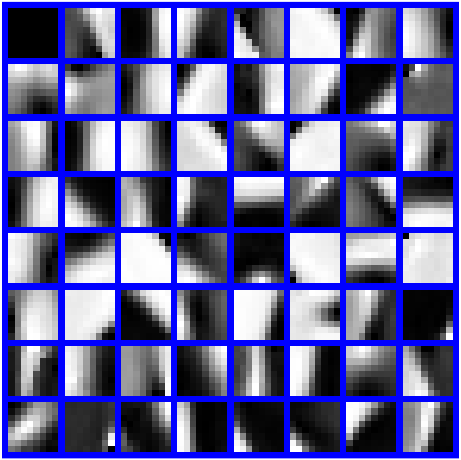}  
  \caption{}
  \label{fig: decomposition D2}
\end{subfigure}

\begin{subfigure}{.19\textwidth}
  \centering
  \includegraphics[width=1\linewidth]{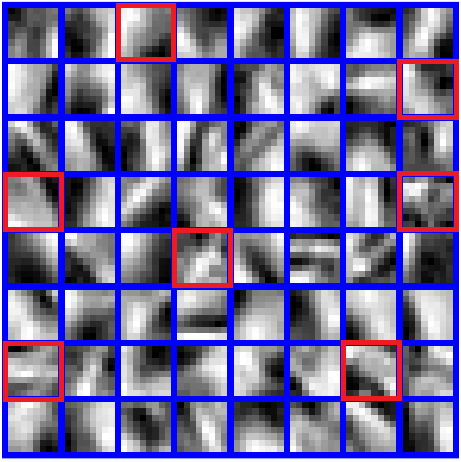}  
  \caption{}
\end{subfigure}
\begin{subfigure}{.19\textwidth}
  \centering
  \includegraphics[width=1\linewidth]{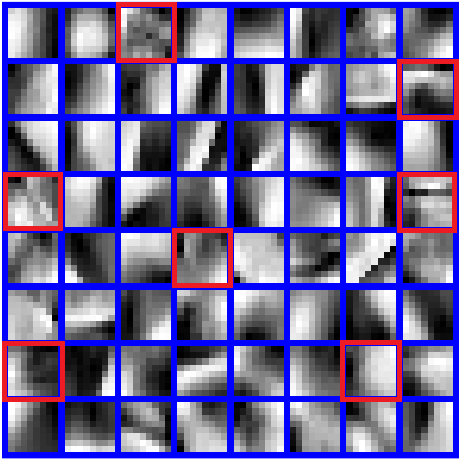}  
  \caption{}
\end{subfigure}
\caption{Example of coupled dictionaries learnt from MR (left) and CT (right) images using the proposed method (a,b) and the standard CDL approach of \cite{CDL2019} (c,d). The red frames indicate weakly correlated atoms.}
\label{fig: dicts with fixed T}
\end{figure}

The proposed CDL method alternates between the sparse coding step using the modified SOMP and the dictionary update using K-SVD. We will refer to the proposed approach as the simultaneous coupled dictionary learning (SCDL). Fig.~\ref{fig: dicts with fixed T} \NO{shows how the proposed SCDL method captures correlated features more efficiently than the standard CDL method of \cite{CDL2019}}. Specifically, one can see how the dictionaries obtained using the standard method (Fig.~\ref{fig: dicts with fixed T}(c,d)) contain atoms that are \NO{relatively uncorrelated} (framed in red), \NO{while all the atoms obtained with SCDL present a clear correlation} (see Fig.~\ref{fig: dicts with fixed T}(a,b)). 

\begin{figure*}[h]
\centering
\begin{subfigure}{.23\textwidth}
  \centering
  \includegraphics[width=1\linewidth]{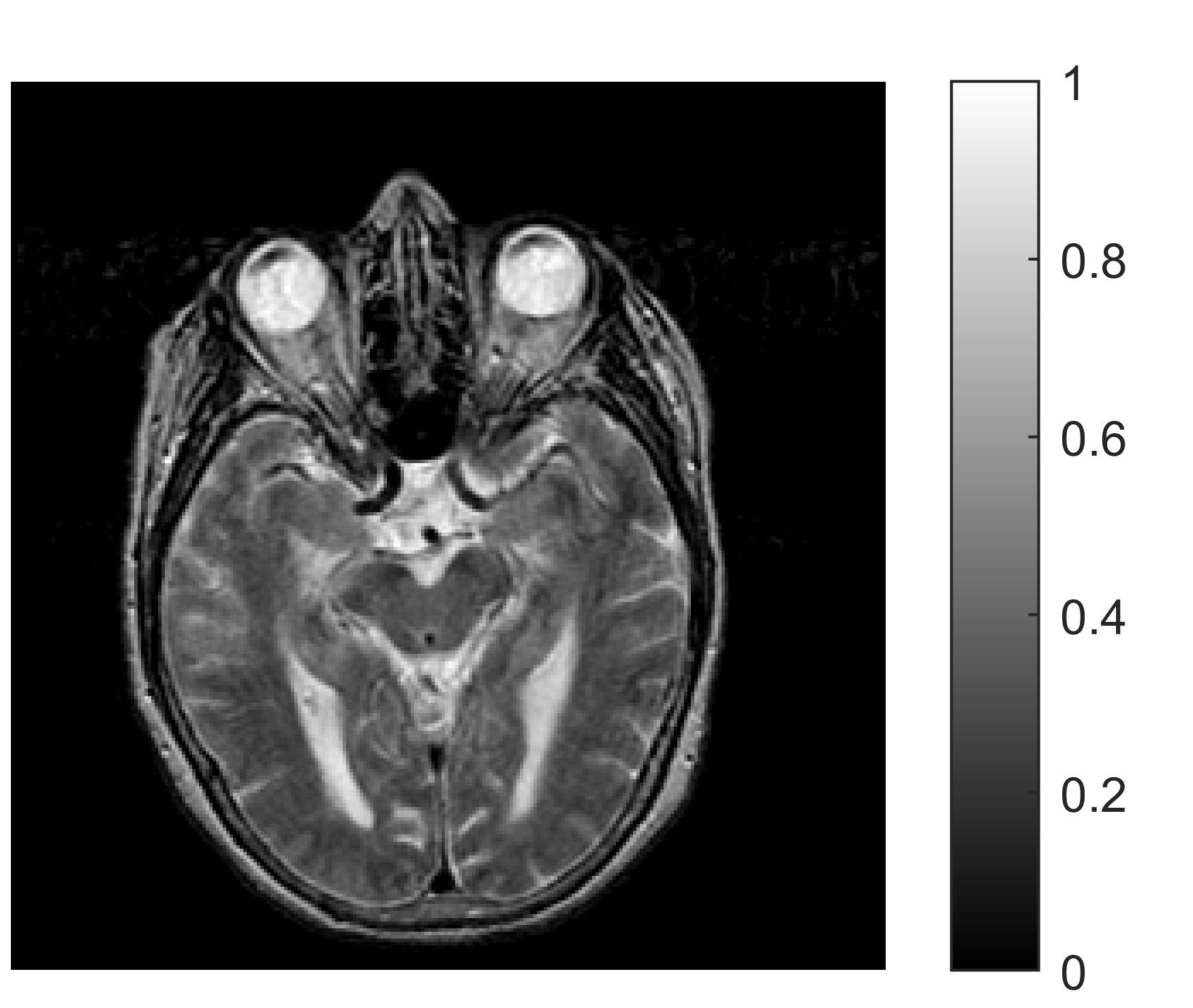}  
  \caption{$\boldsymbol{I}_1$ (MR)}
\end{subfigure}
\begin{subfigure}{.23\textwidth}
  \centering
  \includegraphics[width=1\linewidth]{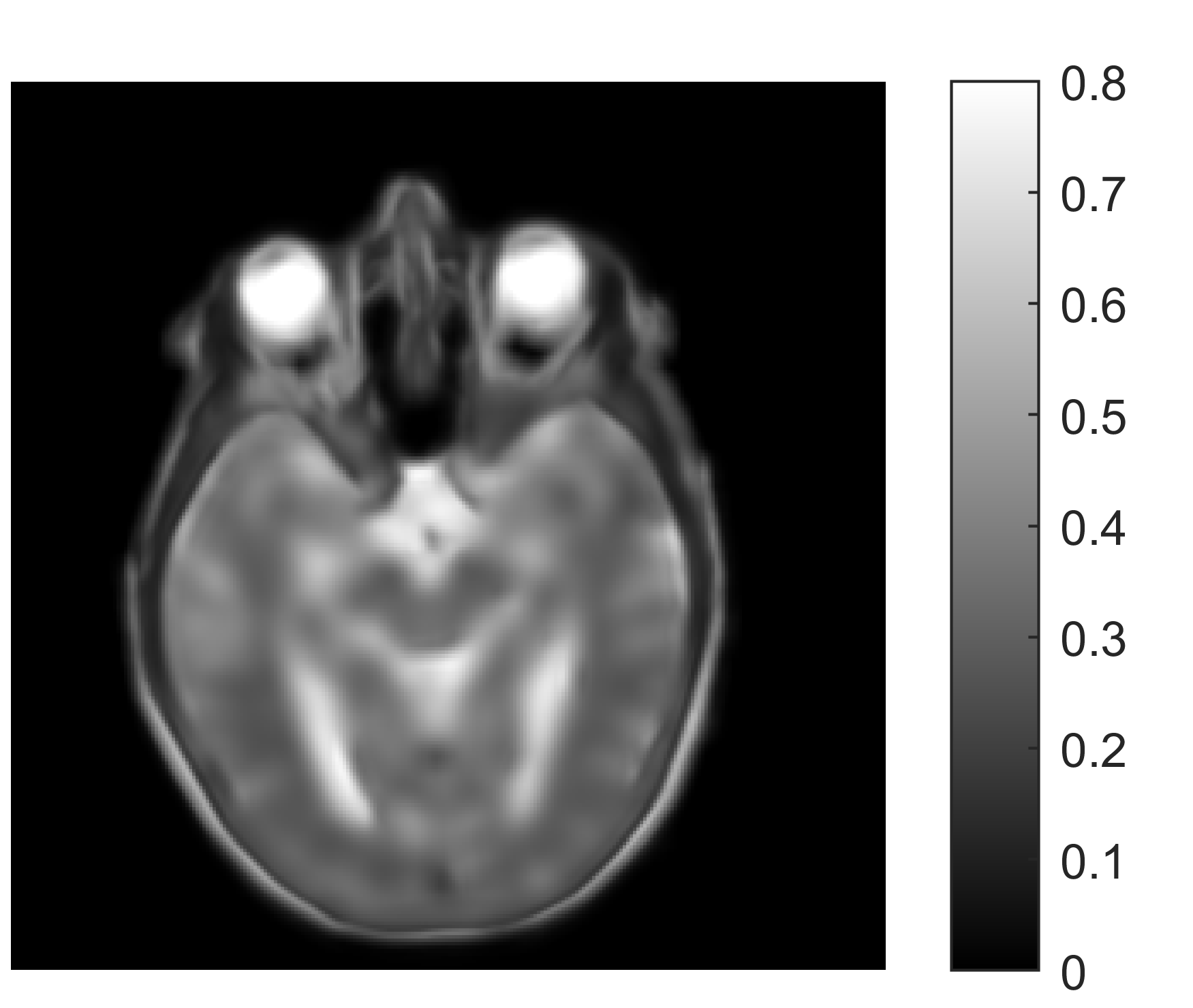}  
  \caption{$\boldsymbol{I}_1^z$}
  \label{fig:sub-z11}
\end{subfigure}
\begin{subfigure}{.23\textwidth}
  \centering
  \includegraphics[width=1\linewidth]{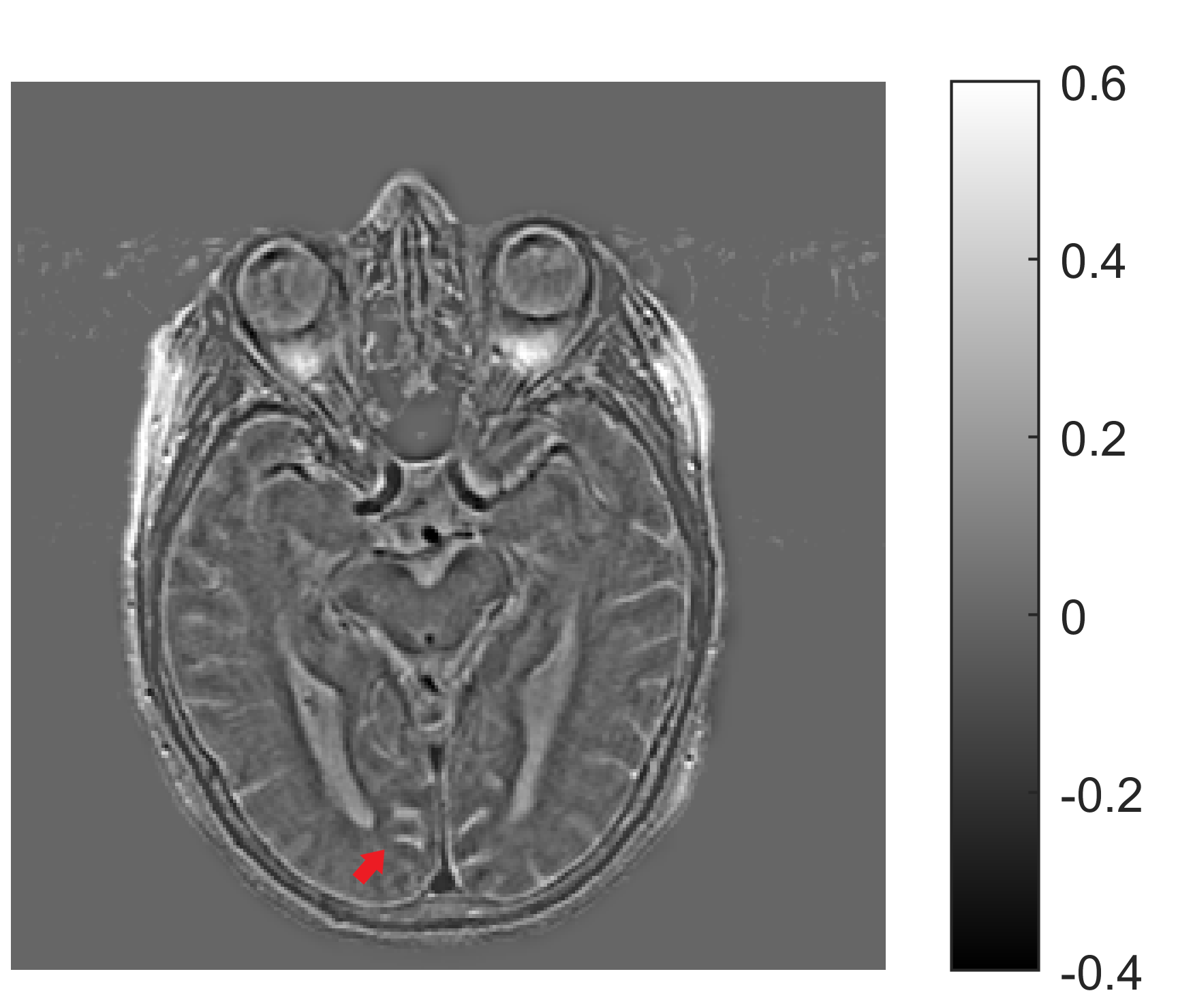}  
  \caption{$\boldsymbol{I}_1^e$}
  \label{fig:sub-e11}
\end{subfigure}
\begin{subfigure}{.23\textwidth}
  \centering
  \includegraphics[width=1\linewidth]{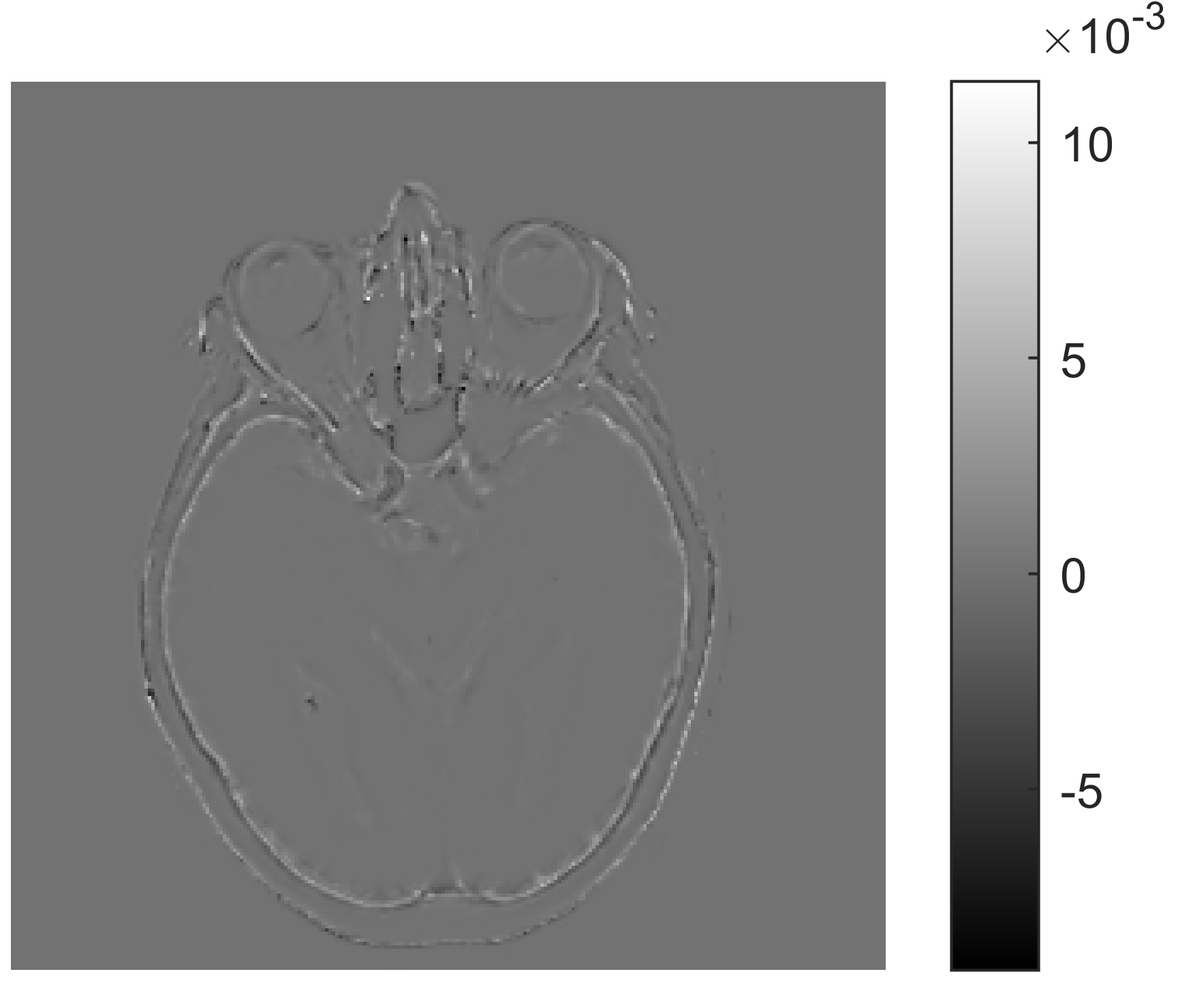}  
  \caption{$\boldsymbol{I}_1-\boldsymbol{I}_1^z-\boldsymbol{I}_1^e$}
  \label{fig:sub-res11}
\end{subfigure}

\begin{subfigure}{.23\textwidth}
  \centering
  \includegraphics[width=1\linewidth]{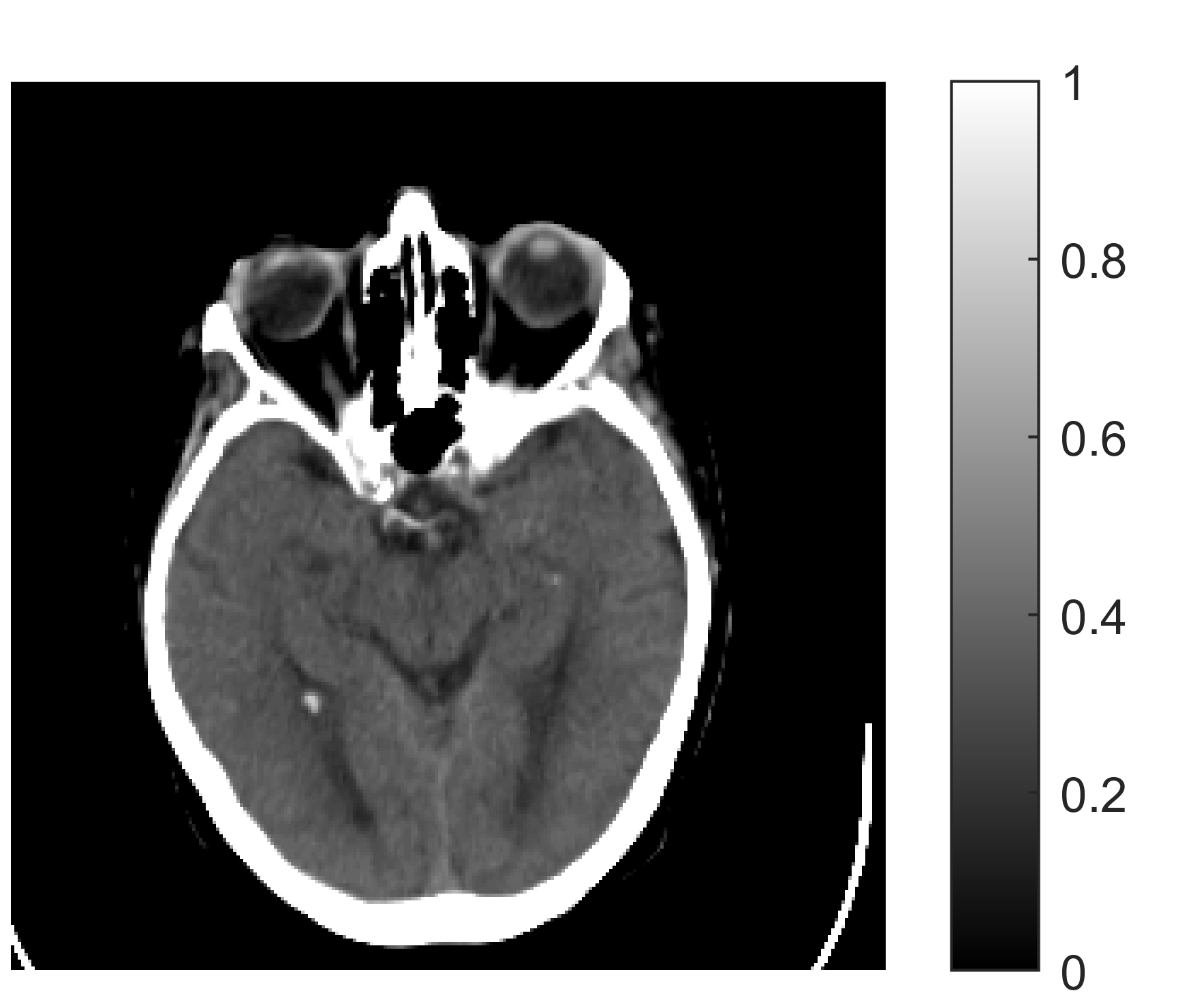}  
  \caption{$\boldsymbol{I}_2$ (CT)}
\end{subfigure}
\begin{subfigure}{.23\textwidth}
  \centering
  \includegraphics[width=1\linewidth]{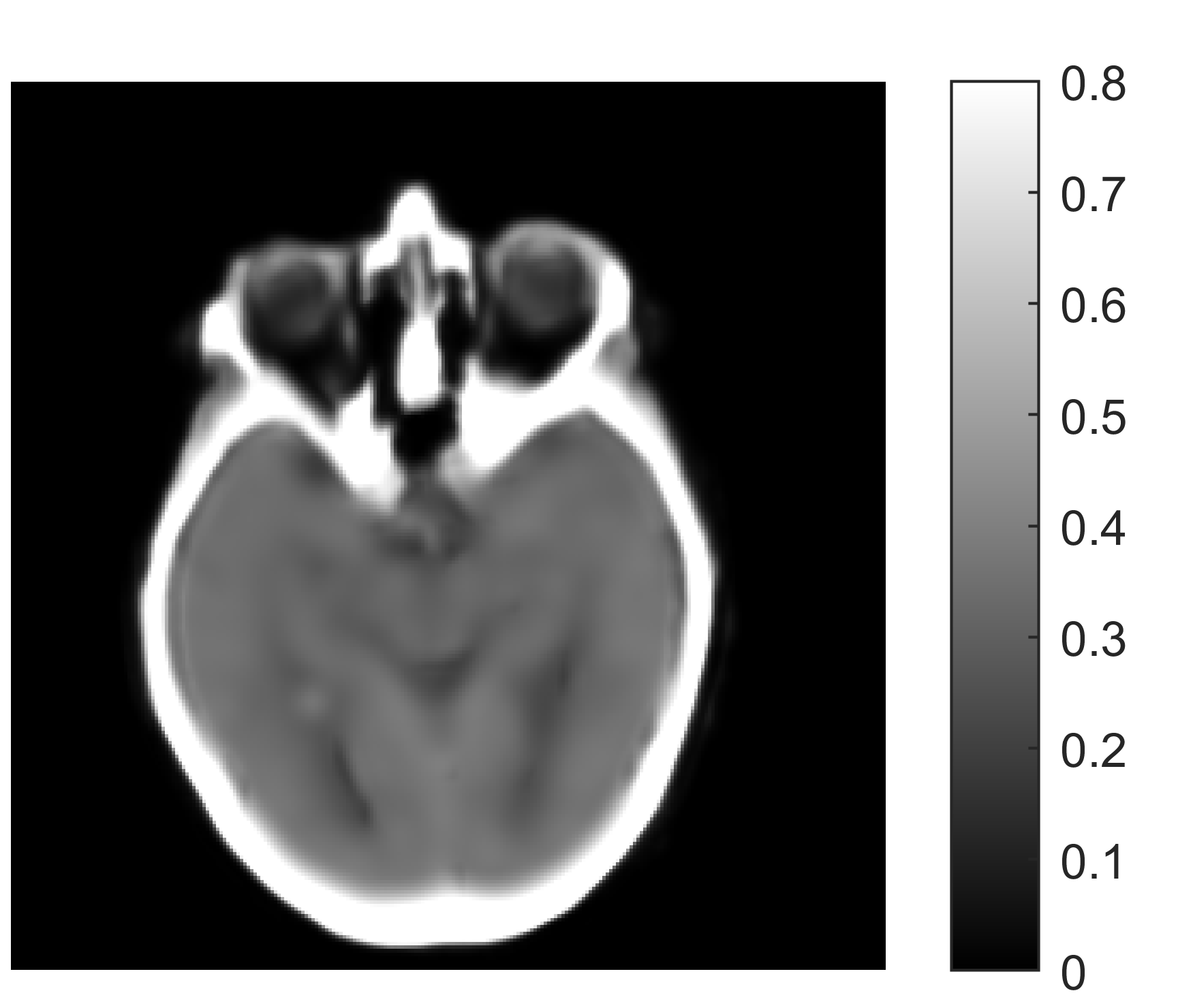}  
  \caption{$\boldsymbol{I}_2^z$}
  \label{fig:sub-z12}
\end{subfigure}
\begin{subfigure}{.23\textwidth}
  \centering
  \includegraphics[width=1\linewidth]{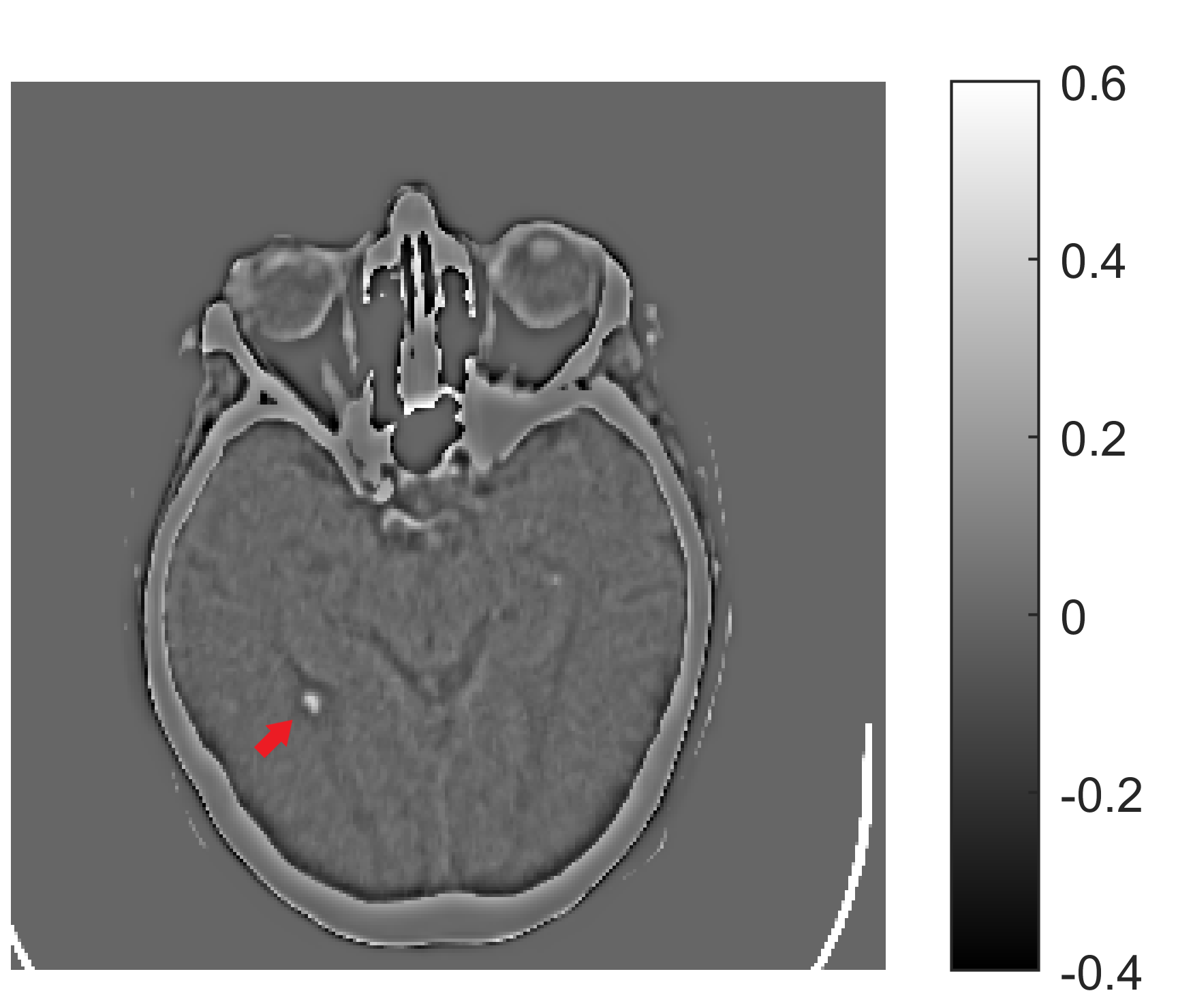}  
  \caption{$\boldsymbol{I}_2^e$}
  \label{fig:sub-e12}
\end{subfigure}
\begin{subfigure}{.23\textwidth}
  \centering
  \includegraphics[width=1\linewidth]{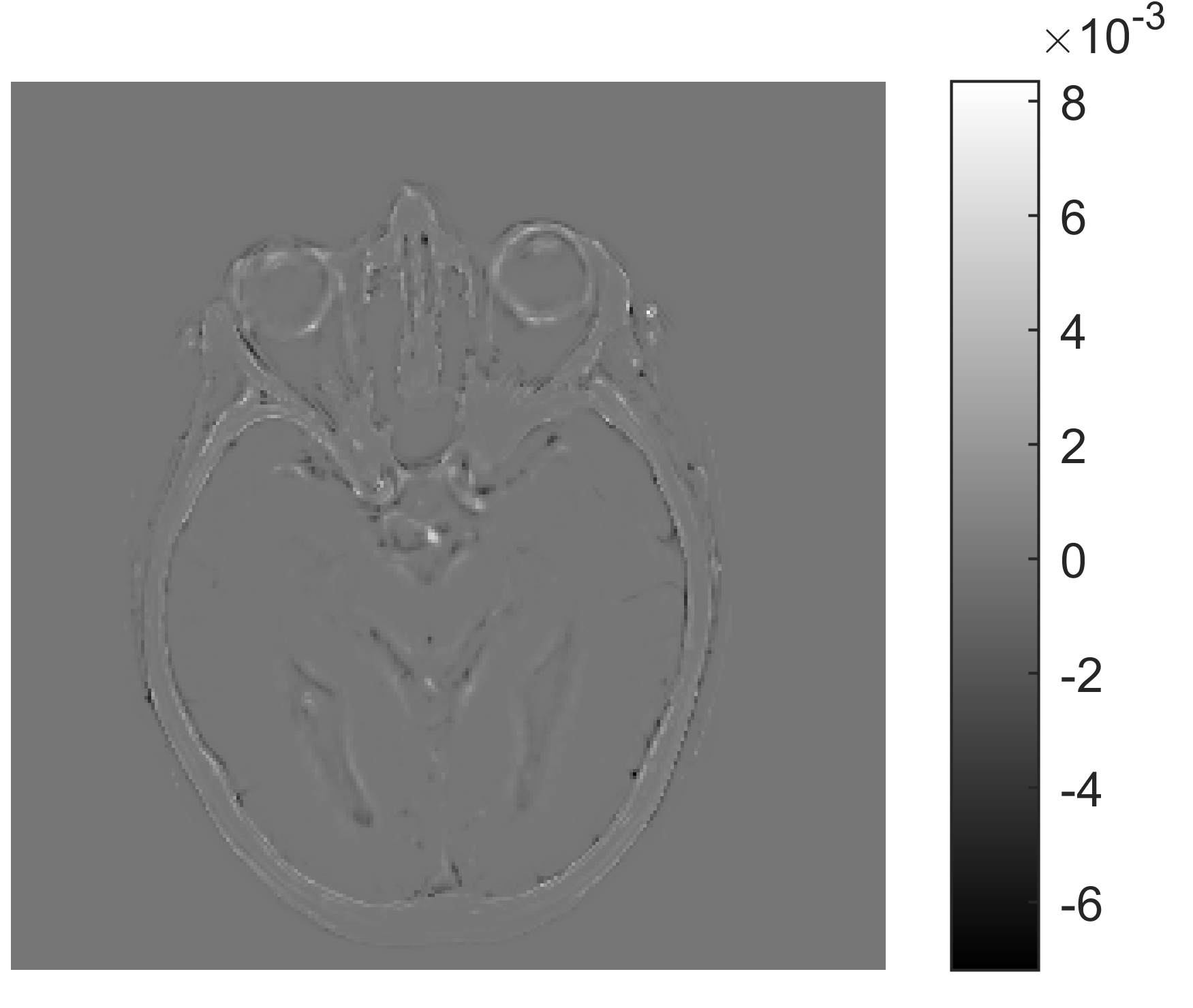}  
  \caption{$\boldsymbol{I}_2-\boldsymbol{I}_2^z-\boldsymbol{I}_2^e$}
  \label{fig:sub-res12}
\end{subfigure}
\caption{A pair of MR and CT input images (a,e) decomposed into their coupled (b,f) and independent (c,g) components based on the proposed model. The residuals are shown in (d,h).}
\label{fig: decomposition-example}
\end{figure*}
\section{Image Decomposition via SCDL}
\label{section_Decomposition}
\NO{Our fusion method relies on the decomposition of the input images into correlated and independent components. The latter represent features that are unique to each modality and that we seek to preserve. This section introduces the proposed decomposition model and explains how to solve the resulting minimization problem.} 
\subsection{Decomposition Model}
\label{subsection: decomposition_model}
\begin{figure*}[]
\centering
\begin{subfigure}{.23\textwidth}
  \centering
  \includegraphics[width=1\linewidth]{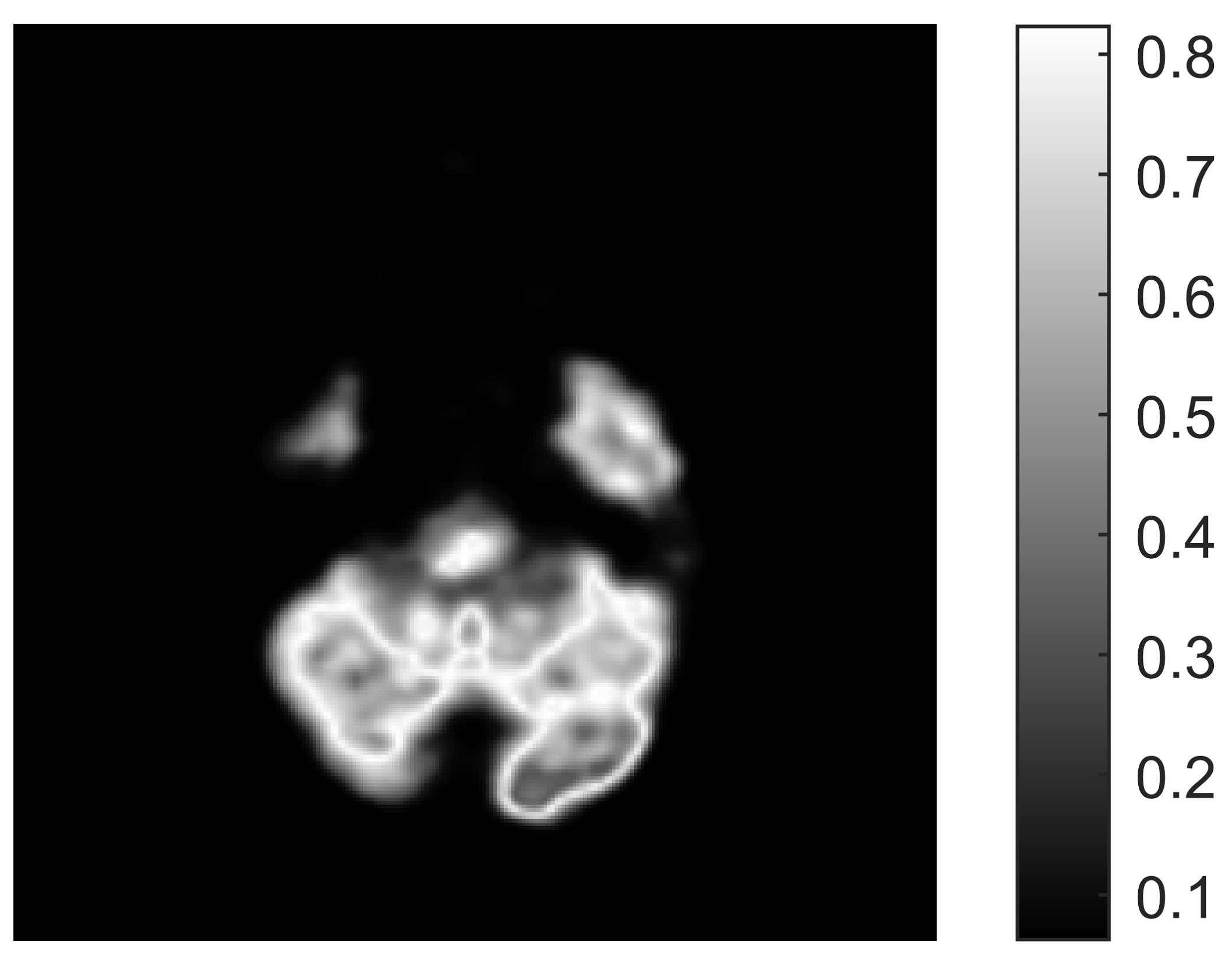}  
  \caption{$\boldsymbol{I}_1$ (PET)}
\end{subfigure}
\begin{subfigure}{.23\textwidth}
  \centering
  \includegraphics[width=1\linewidth]{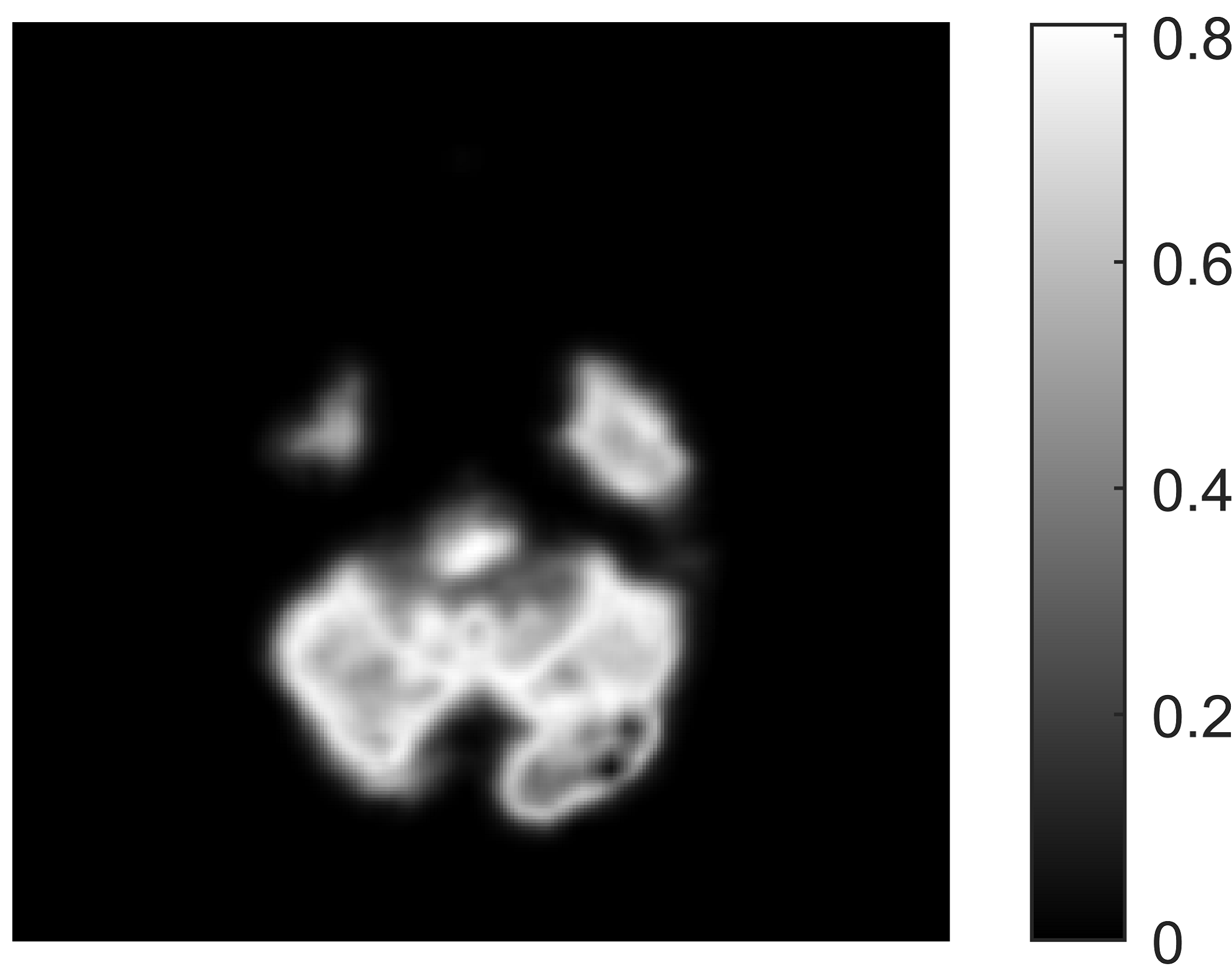}  
  \caption{$\boldsymbol{I}_1^z$}
  \label{fig:sub-z21}
\end{subfigure}
\begin{subfigure}{.23\textwidth}
  \centering
  \includegraphics[width=1\linewidth]{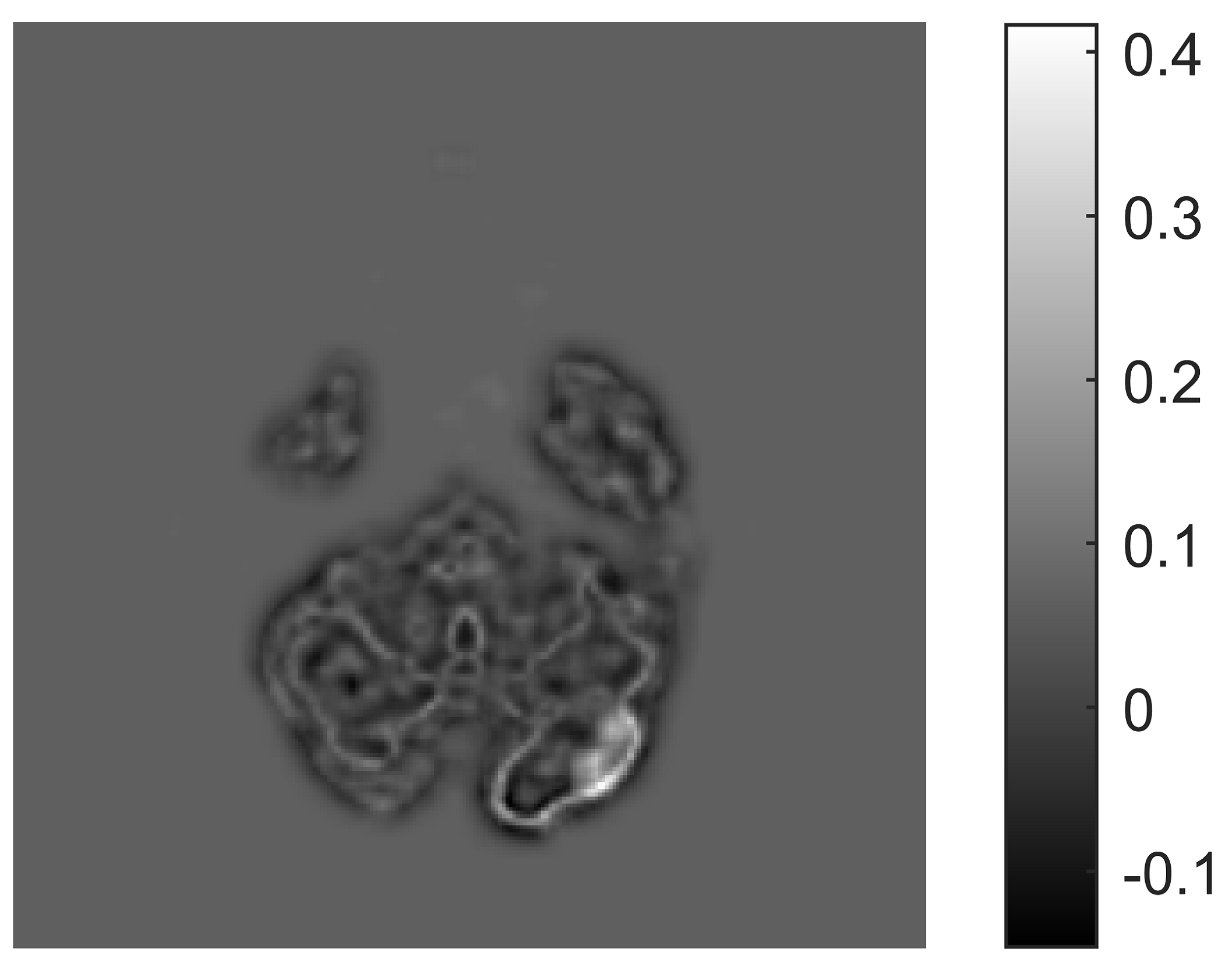}  
  \caption{$\boldsymbol{I}_1^e$}
  \label{fig:sub-e21}
\end{subfigure}
\begin{subfigure}{.23\textwidth}
  \centering
  \includegraphics[width=1\linewidth]{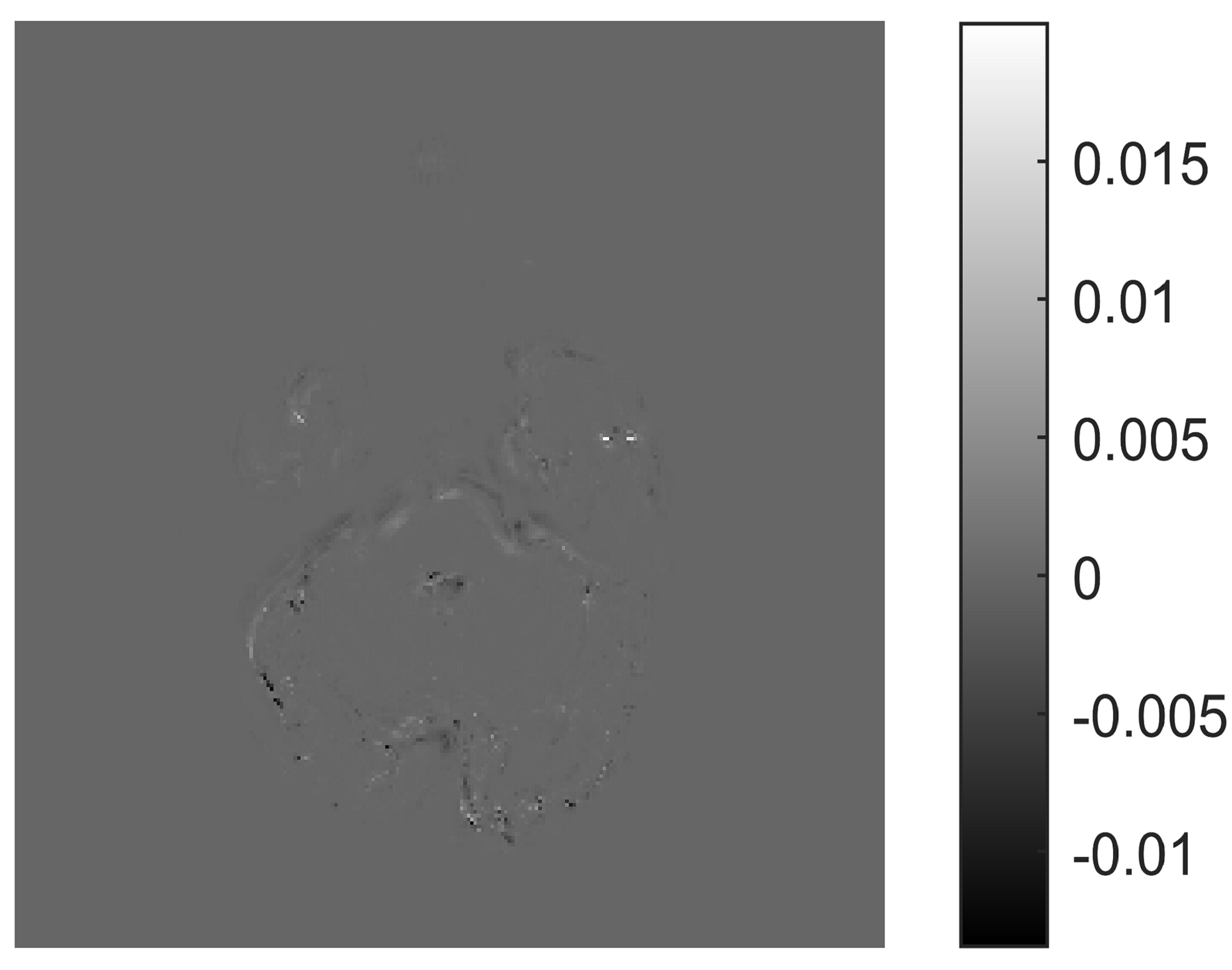}  
  \caption{$\boldsymbol{I}_1-\boldsymbol{I}_1^z-\boldsymbol{I}_1^e$}
  \label{fig:sub-res21}
\end{subfigure}

\begin{subfigure}{.23\textwidth}
  \centering
  \includegraphics[width=1\linewidth]{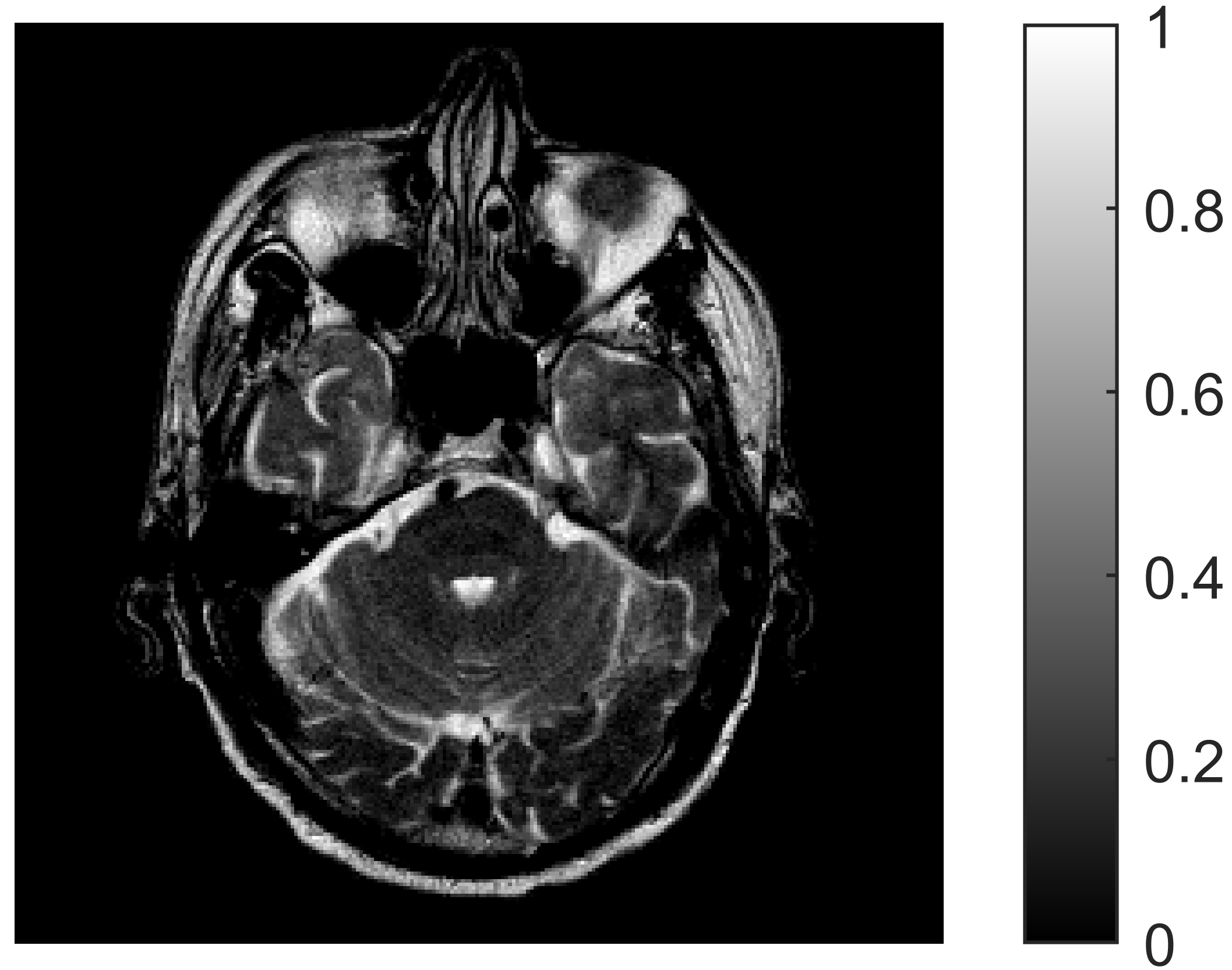}  
  \caption{$\boldsymbol{I}_2$ (MR)}
\end{subfigure}
\begin{subfigure}{.23\textwidth}
  \centering
  \includegraphics[width=1\linewidth]{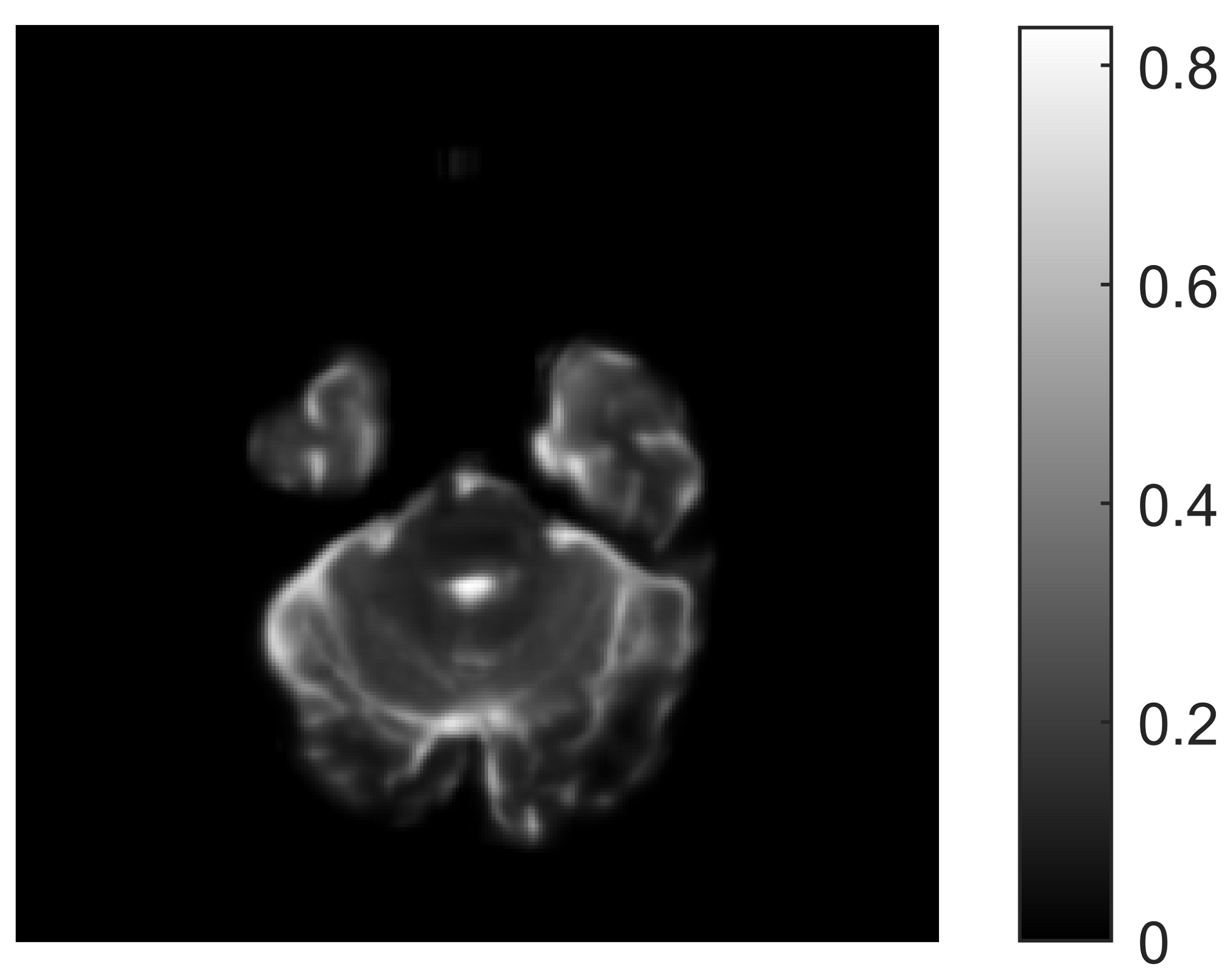}  
  \caption{$\boldsymbol{I}_2^z$}
  \label{fig:sub-z22}
\end{subfigure}
\begin{subfigure}{.23\textwidth}
  \centering
  \includegraphics[width=1\linewidth]{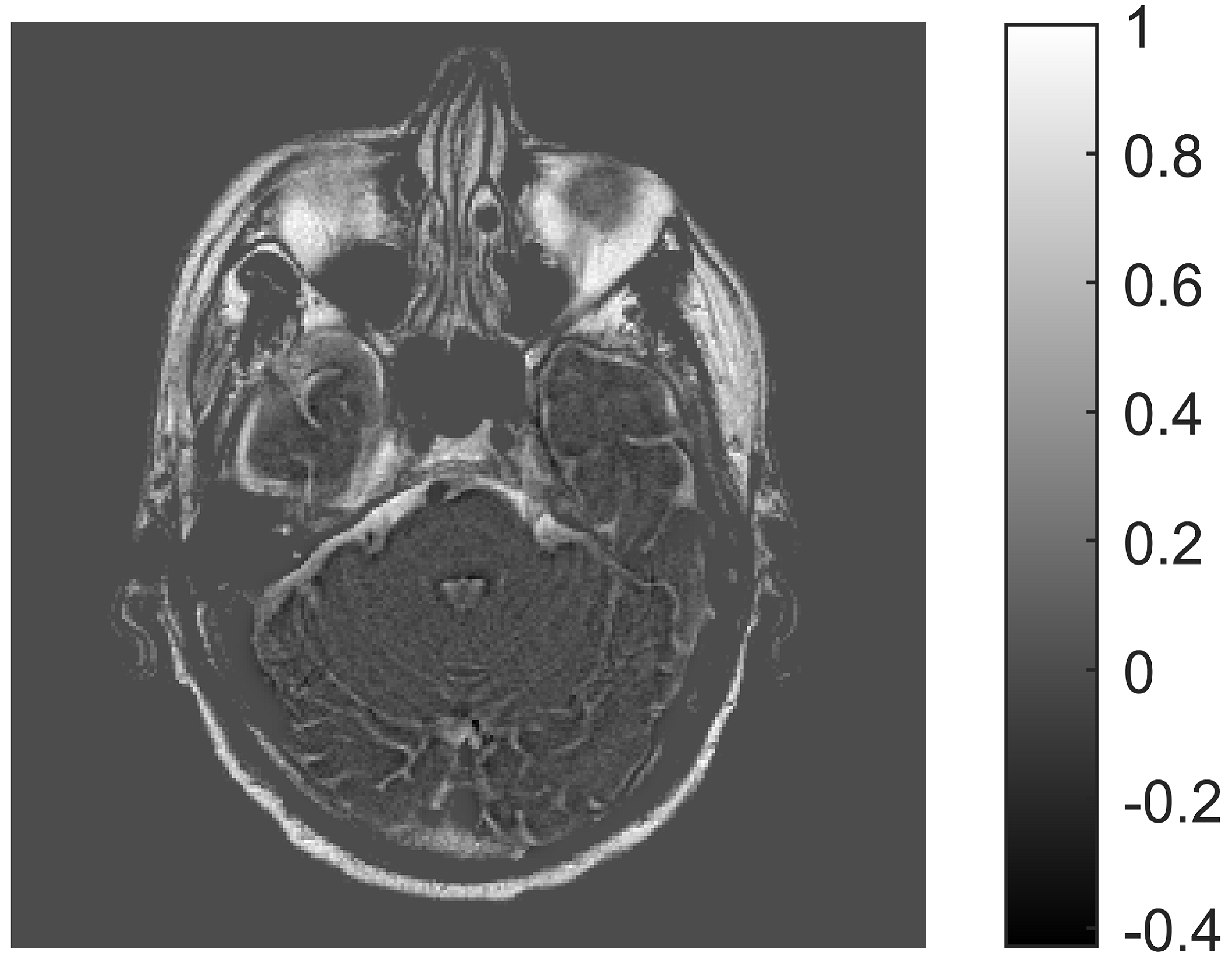}  
  \caption{$\boldsymbol{I}_2^e$}
  \label{fig:sub-e22}
\end{subfigure}
\begin{subfigure}{.23\textwidth}
  \centering
  \includegraphics[width=1\linewidth]{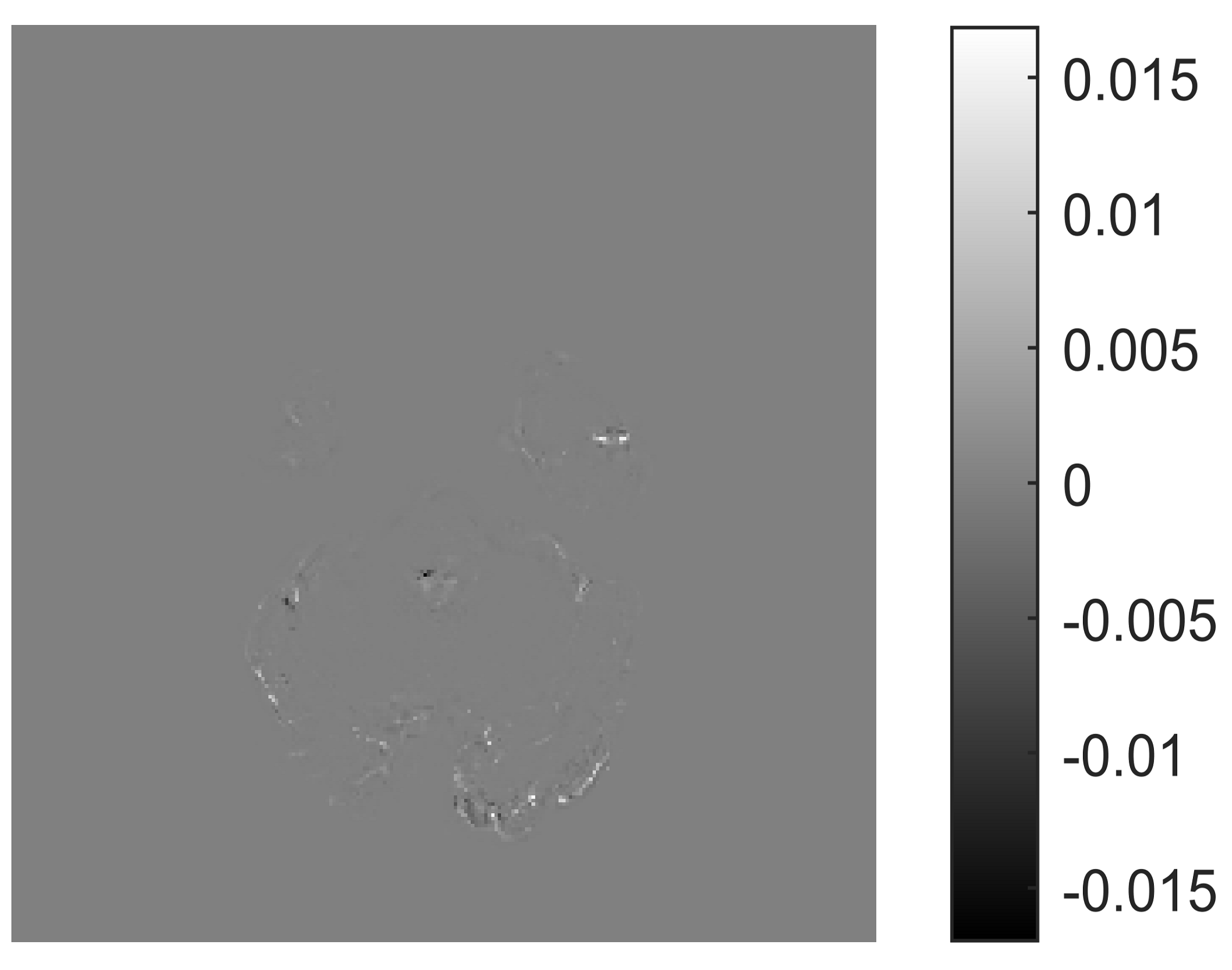}  
  \caption{$\boldsymbol{I}_2-\boldsymbol{I}_2^z-\boldsymbol{I}_2^e$}
  \label{fig:sub-res22}
\end{subfigure}
\caption{A pair of PET and MR input images (a,e) decomposed into their coupled (b,f) and independent (c,g) components using the proposed model. The residuals are shown in (d,h). \hl{(The grey-scale component of the PET image is used in the decomposition, as explained in Subsection~\ref{sub: color-gray fusion}.)}}
\label{fig: decomposition-example2}
\end{figure*}

Let the input images to be fused be denoted by $\boldsymbol{I}_1\in\mathbb{R}^{M\times N}$ and $\boldsymbol{I}_2\in\mathbb{R}^{M\times N}$.\footnote{The images are registered beforehand, which is a typical assumption in the literature\hl{\cite{NSST-PAPCNN2019}}.}
The correlated components are denoted by $\boldsymbol{I}_1^z\in\mathbb{R}^{M\times N}$ and $\boldsymbol{I}_2^z\in\mathbb{R}^{M\times N}$, and the independent components by $\boldsymbol{I}_1^e\in\mathbb{R}^{M\times N}$ and $\boldsymbol{I}_2^e\in\mathbb{R}^{M\times N}$. The proposed decomposition model can be expressed as follows
\begin{equation}
\left\{\begin{array}{lcr}
\boldsymbol{I}_1 = \boldsymbol{I}_1^z+ \boldsymbol{I}_1^e\\
\boldsymbol{I}_2 = \boldsymbol{I}_2^z  + \boldsymbol{I}_2^e \end{array}\right.
\text{where} \  \boldsymbol{I}_1^e\perp \!\!\! \perp \boldsymbol{I}_2^e.
\label{eq: model0}
\end{equation}
\noindent \NO{The SCDL method explained in Section~\ref{section_CDL} operates patch-wise. Therefore, we rewrite \eqref{eq: model0} using the matrices containing all the extracted patches as follows} 
\begin{equation}
\left\{\begin{array}{lcr}
\boldsymbol{X}_1 = \boldsymbol{Z}_1+ \boldsymbol{E}_1\\
\boldsymbol{X}_2 = \boldsymbol{Z}_2  + \boldsymbol{E}_2\end{array}\right.
\text{where} \ \boldsymbol{E}_1\perp \!\!\! \perp \boldsymbol{E}_2,
\label{eq: model}
\end{equation}
\noindent where the matrices $\boldsymbol{X}_1\in \mathbb{R}^{m\times p}$ and $\boldsymbol{X}_2\in \mathbb{R}^{m\times p}$ contain the $p$ vectorized overlapping patches of size $m$ extracted from the input images. The \NO{patches of the} correlated components are represented by $\boldsymbol{Z}_1\in\mathbb{R}^{m\times p}$ and $\boldsymbol{Z}_2\in\mathbb{R}^{m\times p}$, and those of the independent components by $\boldsymbol{E}_1\in\mathbb{R}^{m\times p}$ and $\boldsymbol{E}_2\in\mathbb{R}^{m\times p}$.
Fig.~\ref{fig: decomposition-example} \NO{shows how the proposed decomposition model captures correlated and independent features in a pair of MR-CT images. The independent components contain edges and details that can be clearly observed in only one of the modalities, \textit{e.g.,} sulci details in the MR image and calcification in the CT image (indicated by red arrows). The correlated components represent the underlying joint structure that can be referred to as the base or background layer. Fig.~\ref{fig: decomposition-example2}, shows another example of decomposition for a pair of PET-MR images. In functional-anatomical imaging, any details in the images are naturally independent. The independent components also capture any non-overlapping regions. Finally, the correlated components contain the regions where the background of the anatomical image overlaps with the biological activity information.}
\subsection{\hl{Minimization} Problem}
In order to estimate the correlated and independent components in \eqref{eq: model}, we formulate a minimization problem based on the SCDL approach described in the previous section. Specifically, \NO{we seek the coupled sparse representation of} $\boldsymbol{Z}_1$ and $\boldsymbol{Z}_2$ using sparse codes with identical supports $\boldsymbol{A}_1\in\mathbb{R}^{n\times p}$ and $\boldsymbol{A}_2\in\mathbb{R}^{n\times p}$ and coupled dictionaries $\boldsymbol{D}_1\in\mathbb{R}^{m\times n}$ and $\boldsymbol{D}_2\in\mathbb{R}^{m\times n}$ (\textit{i.e.,} $\boldsymbol{Z}_1 = \boldsymbol{D}_1\boldsymbol{A}_1$ and $\boldsymbol{Z}_2 = \boldsymbol{D}_2\boldsymbol{A}_2$). \NO{The element-wise independence of $\boldsymbol{E}_1$ and $\boldsymbol{E}_2$ is enforced by minimizing the squared Pearson correlation coefficients, where the local means $\mu$ and standard deviations $\sigma$ are estimated patch-wise. The corresponding cost function is expressed as follows}
\[\boldsymbol{\phi}\big([\boldsymbol{E}_1]_{(i,j)},\![\boldsymbol{E}_2]_{(i,j)}\big) \!\!=\!\!\Bigg(\!\frac{([\boldsymbol{E}_1]_{(i,j)} \!-\! \mu_{1,j})([\boldsymbol{E}_2]_{(i,j)}\!-\! \mu_{2,j})}{\sigma_{1,j}\sigma_{2,j}} \!\Bigg)^2\]
\noindent \NO{where $j$ is the patch index and the subscripts $1$ and $2$ indicate the associated components $\boldsymbol{E}_1$ and $\boldsymbol{E}_2$.} \NO{Note that this patch-wise approach results implicitly in multiple counts of the same pixel when considering overlapping patches~\cite{Sulam2014}.
Combining the independence term with the proposed SCDL approach leads to the following minimization problem }
\begin{equation*}
\begin{aligned}
\underset{\substack{\boldsymbol{D}_{1},\boldsymbol{D}_{2},\boldsymbol{A}_{1},\boldsymbol{A}_{2},\\\boldsymbol{E}_{1},\boldsymbol{E}_{2}}}{\mathrm{minimize}}\sum_{\substack{i=1,\dots,m\\j=1,\dots,p}}\boldsymbol{\phi}\big([\boldsymbol{E}_1]_{(i,j)},[\boldsymbol{E}_2]_{(i,j)}\big)
\end{aligned}
\end{equation*}
\begin{equation}
\begin{aligned}
\quad\text{s.t.}& \quad \boldsymbol{D}_1\boldsymbol{A}_1+\boldsymbol{E}_1 = \boldsymbol{X}_1\\
& \quad \boldsymbol{D}_2\boldsymbol{A}_2+\boldsymbol{E}_2 = \boldsymbol{X}_2\\
&\quad \operatorname{supp}\{\boldsymbol{A}_1\} = \operatorname{supp}\{\boldsymbol{A}_2\}\\
&  \quad\left\| [\boldsymbol{A}_1]_i \right\| _0 \leq T, \left\| [\boldsymbol{A}_2]_i \right\| _0 \leq T, \forall i\\
&\quad\left\| [\boldsymbol{D}_1]_t \right\|_{2}= 1, \left\| [\boldsymbol{D}_2]_t \right\|_{2}= 1, \forall t,
\end{aligned} \label{eq:obj_fct}
\end{equation}
\noindent \NO{where the sparsity and common support constraints have been introduced in \eqref{cor_estimation_CDL}.} 
\subsection{\hl{Optimization}}
\NO{Problem \eqref{eq:obj_fct} is challenging because of its non-convexity and the presence of multiple sets of variables. Therefore, we propose to break the optimization procedure into simpler subproblems where we consider minimization with respect to separate blocks of variables. We then alternate between these subproblems until finding a local optimum. Specifically, the minimization with respect to the sparse codes and dictionaries, and the minimization with respect to the independent components are treated separately. Furthermore, we simplify the problem by approximating the first two equality constraints in \eqref{eq:obj_fct} by quadratic approximation terms. This leads to a new optimization problem that can be written as}
\begin{multline*}
\underset{\substack{\boldsymbol{D}_{1},\boldsymbol{D}_{2},\boldsymbol{A}_{1},\boldsymbol{A}_{2},\\\boldsymbol{E}_{1},\boldsymbol{E}_{2}}}
{\mathrm{minimize}}\sum_{\substack{i=1,\dots,m\\j=1,\dots,p}}\boldsymbol{\phi}\big([\boldsymbol{E}_1]_{(i,j)},[\boldsymbol{E}_2]_{(i,j)}\big)\\ +\frac{\rho}{2}\sum_{k=1,2}\| \boldsymbol{D}_k\boldsymbol{A}_k+\boldsymbol{E}_k-\boldsymbol{X}_k\|_F^2
\end{multline*}
\begin{equation}
\begin{aligned}
\text{s.t.} \quad &\operatorname{supp}\{\boldsymbol{A}_1\} = \operatorname{supp}\{\boldsymbol{A}_2\}\\
 \qquad &\left\| [\boldsymbol{A}_1]_i \right\| _0 \leq T, \left\| [\boldsymbol{A}_2]_i \right\| _0 \leq T, \forall i\\
 \qquad &\left\| [\boldsymbol{D}_1]_t \right\|_{2}= 1, \left\| [\boldsymbol{D}_2]_t \right\|_{2}= 1, \forall t
\end{aligned}\label{eq: relaxed opt. problem}
\end{equation}\\
\noindent where $\rho>0$ \NO{controls the} trade-off between \NO{the independence of $\boldsymbol{E}_1$ and $\boldsymbol{E}_2$ and the accuracy of the sparse representations}. \NO{As mentioned above, problem \eqref{eq: relaxed opt. problem} is tackled by alternating minimizations with respect to the two blocks of variables $\{\boldsymbol{A_{1}},\boldsymbol{A_{2}},\boldsymbol{D_{1}},\boldsymbol{D_{2}}\}$ and $\{\boldsymbol{E_{1},\boldsymbol{E_{2}}}\}$. Each resulting subproblem is described in more detail in the following.}

    \subsubsection{Optimization with respect to $\{\boldsymbol{A_{1}},\boldsymbol{A_{2}},\boldsymbol{D_{1}},\boldsymbol{D_{2}}\}$}
    \NO{The first optimization subproblem can be written as}
\begin{equation*}
\begin{split}
\begin{aligned}
&\underset{{\boldsymbol{D}_1,\boldsymbol{D}_2, \boldsymbol{A}_1, \boldsymbol{A}_2}}{\mathrm{minimize}} \| \boldsymbol{D}_1\boldsymbol{A}_1-\boldsymbol{X}_1^{'}\|_F^2 + \| \boldsymbol{D}_2\boldsymbol{A}_2-\boldsymbol{X}_2^{'}\|_F^2\\
&\text{s.t.} \quad \operatorname{supp}\{\boldsymbol{A}_1\} = \operatorname{supp}\{\boldsymbol{A}_2\}\\
& \qquad \left\| [\boldsymbol{A}_1]_i \right\| _0 \leq T, \left\| [\boldsymbol{A}_2]_i \right\| _0 \leq T, \forall i\\
& \qquad \left\| [\boldsymbol{D}_1]_t \right\|_{2}= 1, \left\| [\boldsymbol{D}_2]_t \right\|_{2}= 1, \forall t 
\end{aligned}\label{eq: CDL step}
\end{split}
\end{equation*}
where $ \boldsymbol{X}_1^{'} = \boldsymbol{X}_1 - \boldsymbol{E}_1$ and $ \boldsymbol{X}_2^{'} = \boldsymbol{X}_2 - \boldsymbol{E}_2$. This \NO{subproblem} is addressed using the SCDL method explained in Section~\ref{section_CDL}. \NO{The dictionaries can be initialized in the first iteration of the algorithm using a predefined dictionary, \textit{e.g.,} based on discrete cosine transforms (DCT). In subsequent iterations, the SCDL method is initialized using the dictionaries obtained from the previous one. } 
\NO{Furthermore, initializing the sparsity parameter $T$ with a small value and gradually increasing it at each iteration ensures a warm start of the algorithm.}
%
\subsubsection{Optimization with respect to $\{\boldsymbol{E}_1,\boldsymbol{E}_2\}$}
The second optimization subproblem can be written as
\begin{multline}
\underset{\boldsymbol{E}_1,\boldsymbol{E}_2}{\mathrm{minimize}} \sum_{\substack{i=1,\dots,m\\j=1,\dots,p}}\boldsymbol{\phi}\big([\boldsymbol{E}_1]_{(i,j)},[\boldsymbol{E}_2]_{(i,j)}\big)\\ + \frac{\rho}{2}\sum_{k=1,2}\| \boldsymbol{D}_k\boldsymbol{A}_k+\boldsymbol{E}_k- \boldsymbol{X}_k\|_F^2.
\label{eq: EM step}
\end{multline}
\noindent The \NO{estimates of $\{\bb{E}_1,\bb{E}_2\}$ depend on the patch-wise means and standard deviations, which in turn depend on the independent components themselves. Therefore, we propose to address this subproblem using an expectation-maximization (EM) method}. 
\NO{The EM approach leads to the following updates for the independent components} 
\begin{equation*}
\begin{aligned}\label{eq: E update}
[\boldsymbol{E}_1]_{(i,j)}^{+}\! = \!\frac{\rho[{\scriptstyle\boldsymbol{X}_1-\boldsymbol{D}_1\boldsymbol{A}_1}]_{(i,j)}\!+\!2\frac{([\boldsymbol{E}_2]_{(i,j)}-{\mu}_{2,j})^2}{\mathrm{max}({\sigma}_{1,j}^2{\sigma}_{2,j}^2,\delta)}{\mu}_{1,j}}{\rho+2\frac{([\boldsymbol{E}_2]_{(i,j)}-{\mu}_{2,j})^2}{\mathrm{max}({\sigma}_{1,j}^2{\sigma}_{2,j}^2,\delta)}}\\
[\boldsymbol{E}_2]_{(i,j)}^{+}\! = \!\frac{\rho[{\scriptstyle\boldsymbol{X}_2-\boldsymbol{D}_2\boldsymbol{A}_2}]_{(i,j)}\!+\!2\frac{([\boldsymbol{E}_1]_{(i,j)}-{\mu}_{1,j})^2}{\mathrm{max}({\sigma}_{1,j}^2{\sigma}_{2,j}^2,\delta)}{\mu}_{2,j}}{\rho+2\frac{([\boldsymbol{E}_1]_{(i,j)}-{\mu}_{1,j})^2}{\mathrm{max}({\sigma}_{1,j}^2{\sigma}_{2,j}^2,\delta)}}
\end{aligned}
\end{equation*}
\noindent where the means and standard deviations are \NO{computed using the} current values of $\{\boldsymbol{E_1},\boldsymbol{E_2}\}$, with $\delta>0$ a constant used to avoid division by zero. \NO{Note that $\{\boldsymbol{E}_1,\boldsymbol{E}_2\}$ can be initialized with zero matrices.}
%
\subsection{Computational Complexity}
\NO{The computational cost of the proposed decomposition algorithm is dominated by the first subproblem (solved using SCDL).  In particular, the complexity of one SCDL iteration (including all substeps) is $\mathcal{O}(p\max(Tnm,T^2m,T^3,nm^2))$, while the complexity of the EM step is only $\mathcal{O}(mp)$. Based on the experimental findings of \cite{CSC2016} and \cite{CSMCA2019}, sparse approximations are performed in this work in single precision, which improves the computational efficiency of the algorithm.}
%
\section{Multimodal Fusion \NO{Rule}}
\label{section_fusion} 
\NO{Once the correlated and independent components are estimated, the final fused image is obtained using an appropriate fusion rule.  In this work, the different components are handled separately. Specifically, we assume that the correlated components should be fused because they contain redundant information, \textit{i.e.}, shared underlying structures in anatomical imaging or background elements in the functional-anatomical case. In contrast, the independent components should be entirely preserved in the final image to avoid loss of modality-specific information (\textit{e.g.}, the calcification inside the CT image of Fig.~\ref{fig: dicts with fixed T}-(g)).} 
\subsection{Fusion of Correlated Components}
\NO{Based on the assumptions mentioned above, the correlated components are fused using a binary selection where the most relevant features are chosen based on the magnitudes of the sparse coefficients. Recall that the proposed SCDL method with common supports allows correlated features to be captured with varying significance levels for each modality. This is a novel approach compared to the standard max-absolute-value rule with a single predefined basis~\cite{SRMCA2014,SR-SOMP2012}. Precisely, the most significant features are selected based on the sparse coefficients with the largest magnitudes as follows} 
\begin{equation*}
\begin{split}
 [\boldsymbol{A}_1]_{(i,j)}^{'} =& \begin{cases}
    [\boldsymbol{A}_1]_{(i,j)}, & \text{if $|[\boldsymbol{A}_1]_{(i,j)}|\geq|[\boldsymbol{A}_2]_{(i,j)}|$}\\
    0, & \text{otherwise}
  \end{cases}, \\
 [\boldsymbol{A}_2]_{(i,j)}^{'} =& \begin{cases}
    [\boldsymbol{A}_2]_{(i,j)}, & \text{if $|[\boldsymbol{A}_2]_{(i,j)}|>|[\boldsymbol{A}_1]_{(i,j)}|$}\\
    0, & \text{otherwise}
  \end{cases}.
\end{split}
\end{equation*}
\NO{Then, the fused correlated component, denoted by $\bb{Z}_F$, is reconstructed using the selected coefficients as follows}  
\begin{equation}
\begin{split}
\boldsymbol{Z}_F = \left[ \boldsymbol{D}_1 \  \boldsymbol{D}_2\right]
\begin{bmatrix}
\boldsymbol{A}_1^{'}\\
\boldsymbol{A}_2^{'}\end{bmatrix}.\\
\end{split}\label{eq: fuse_dependents}
\end{equation}
\noindent 
%
\subsection{Final Fused Image}
\NO{Since the independent components contain details or non-overlapping regions that should be preserved in the final image, they can be added directly to the fused correlated components, \textit{i.e.,}}
\begin{equation*}
\boldsymbol{X}_F = \boldsymbol{Z}_F + \boldsymbol{E}_1 + \boldsymbol{E}_2. \label{eq: fusion}
\end{equation*}
\noindent Finally, the fused image is reconstructed by placing the patches of $\boldsymbol{X}_F$ at their \NO{original} positions \NO{in the image} and averaging the overlapping \NO{ones}. Note that the decomposition residuals (see examples in Figs.~\ref{fig:sub-res11}, \ref{fig:sub-res12}, \ref{fig:sub-res21} and \ref{fig:sub-res22}) are considered as noise and not included in the final fused image. \begin{figure}[!h]
\centering
\begin{subfigure}{.23\textwidth}
  \centering
  \includegraphics[width=1\linewidth]{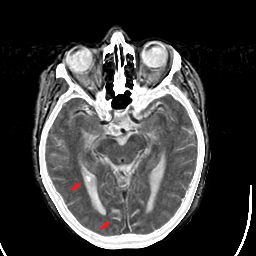} 
  \caption{}
  \label{fig: decomposition D1}
\end{subfigure}
\begin{subfigure}{.23\textwidth}
  \centering
  \includegraphics[width=1\linewidth]{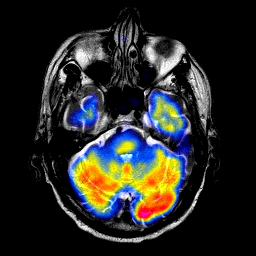}  
  \caption{}
  \label{fig: decomposition D2}
\end{subfigure}
\caption{Final fused images obtained with the proposed method for the MR-CT example in Fig.~\ref{fig: decomposition-example} (a), and MR-PET example in Fig.~\ref{fig: decomposition-example2}. Red arrows indicate the calcification and sulci details preserved by the proposed method.}
\label{fig: fusion res}
\end{figure}
%
%

\hl{Fig.~\ref{fig: fusion res} shows the fused image obtained with the proposed method for the MR-CT images in Fig.~\ref{fig: decomposition-example} and the MR-PET images in Fig.~\ref{fig: decomposition-example2}. The fused MR-CT image contains the modality-specific information captured by the independent components (\textit{e.g.,} calcification and sulci details), as well as the most visible features selected from the correlated components. The fused MR-PET image combines the background of the MR image (Fig.~\ref{fig:sub-z21}) with the functional information from the PET image at the overlapping regions. The details and non-overlapping regions, captured in the independent components appear unaltered in the fused image.}

\subsection{\NO{Color Images}}\label{sub: color-gray fusion}
Functional images are usually displayed in a color-code, \NO{as opposed to the grey-scale anatomical ones}. \NO{One common approach for dealing with the fusion of color images is to} convert them from the original RGB format to the YCbCr (or YUV) color-space~\cite{NSST-PAPCNN2019,CNN2017}. In this new color-space, component Y (\textit{i.e.,} luminance) provides the grey-scale version of the image, which is used here for fusion. \NO{Since the full color information comes from the functional images, the remaining color components (\textit{i.e.}, Cb and Cr) are transmitted directly to the final (grey-scale) fused image.}
\section{Experiments}\label{section_results}
In this section, the proposed multimodal image fusion method \NO{is compared to state-of-the-art fusion methods. The comparison is conducted} in terms of objective metrics, visual quality, and computational efficiency. \NO{The influence of different parameters is also discussed.}
\subsection{\NO{Experiment} Setup}
\subsubsection{\NO{Datasets}} 
We use multimodal images from The Whole Brain Atlas database~\cite{atlas2020} for our experiments. All of the images are registered and are of size $256\times 256$ pixels. Six datasets of \NO{multimodal} MR(T2)--CT, MR(T1)--MR(T2), MR(T2)--PET, MR(Gad)--PET, MR(T2)--SPECT(TC) and MR(T2)--SPECT(TI) images \NO{are used for testing}. Each dataset includes $10$ pairs of images. 
\subsubsection{\NO{Other} Methods}
The proposed fusion method is compared to five \NO{recent} medical image fusion methods, including (1) PCNN-NSST~\cite{NSST-PAPCNN2019} and (2) LLF-IOI~\cite{LLF-IOI2017} (introduced in Section \ref{section_introduction}); (3) LP-CNN~\cite{CNN2017}, which relies on convolutional neural networks and Laplacian pyramids;  (4) a method using convolutional sparse coding referred to as CSR~\cite{CSR2016}; (5) a method using union Laplacian pyramids referred to as ULAP~\cite{ULAP2016}. \NO{All methods considered for comparison are implemented using MATLAB}. All experiments are performed on a PC running an Intel(R) Core(TM) i5-8365U 1.60GHz CPU. %
\NO{Note that} LLF-IOI \NO{is} an anatomical-functional image fusion method. \NO{Therefore, it is tested in our experiments for this type of data only}. \NO{To ensure a fair comparison in the case of anatomical-functional images,} the grey-scale CSR method is adopted to color images using the approach explained in Section~\ref{sub: color-gray fusion}.
\subsubsection{Parameter Setting}\label{Parameters setting}
\NO{The parameters of the proposed decomposition method are tuned to provide the best overall performance with respect to the objective metrics used for evaluation. The best parameters are} $m=64$ \NO{(fully overlapping patches of size $8\times8$)}, $n=128$, $5$ iterations of the decomposition algorithm, $T = 5$, $\rho=10$, $\epsilon = 10^{-4}$, and $\delta=10^{-7}$. 
%
\NO{For all other methods, the parameters are set as indicated in the corresponding papers.} \NO{In section~\ref{section_parameffect}, we explain in more detail the tuning of the parameters and discuss their effect on the fusion performance.}
\subsubsection{Objective Metrics} \NO{The quantitative comparison of the methods is performed based on the following metrics:} the tone-mapped image quality, which measures preservation of intensity and structural information ($TMQI$)\cite{TMQI}, the similarity based fusion quality metric $Q_Y$\cite{QY}, the human visual system-based metric $Q_{CB}$\cite{QCB}, and standard deviation (STD). \NO{Note that high $TMQI$ values correspond to high fidelity in terms of intensity and structural features, high $Q_Y$ values indicate high structural similarities, high $Q_{CB}$ values indicate good visual compatibility, and high $STD$s imply an improved contrast.}
\begin{figure*}[h]
\centering
\begin{subfigure}{.13\textwidth}
  \centering
  \includegraphics[width=1\linewidth]{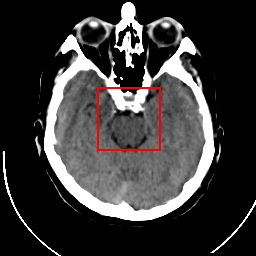}  
\end{subfigure}
\begin{subfigure}{.13\textwidth}
  \centering
  \includegraphics[width=1\linewidth]{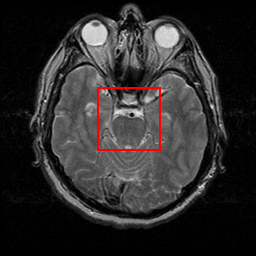}  
\end{subfigure}
\begin{subfigure}{.13\textwidth}
  \centering
  \includegraphics[width=1\linewidth]{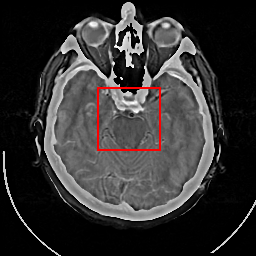}
\end{subfigure}
\begin{subfigure}{.13\textwidth}
  \centering
  \includegraphics[width=1\linewidth]{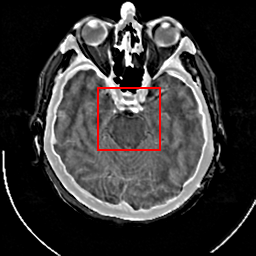}
\end{subfigure}
\begin{subfigure}{.13\textwidth}
  \centering
  \includegraphics[width=1\linewidth]{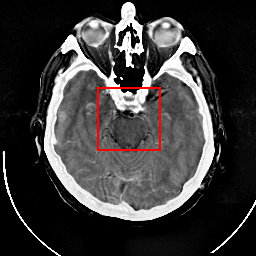}
\end{subfigure}
\begin{subfigure}{.13\textwidth}
  \centering
  \includegraphics[width=1\linewidth]{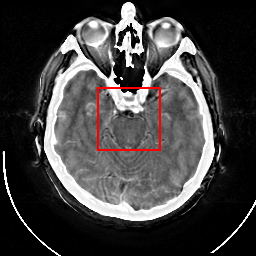}
\end{subfigure}
\begin{subfigure}{.13\textwidth}
  \centering
  \includegraphics[width=1\linewidth]{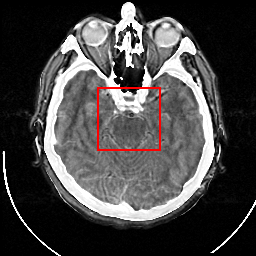}
\end{subfigure}

\vspace{1mm}

\begin{subfigure}{.13\textwidth}
  \centering
  \includegraphics[width=1\linewidth]{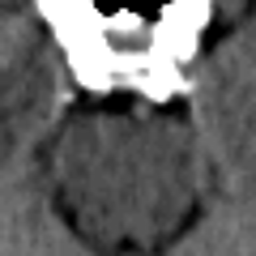}  
  \caption{CT}
\end{subfigure}
\begin{subfigure}{.13\textwidth}
  \centering
  \includegraphics[width=1\linewidth]{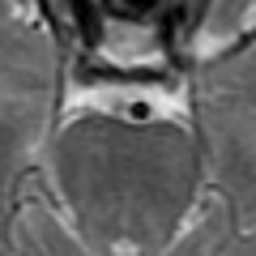}  
  \caption{MR(T2)}
\end{subfigure}
\begin{subfigure}{.13\textwidth}
  \centering
  \includegraphics[width=1\linewidth]{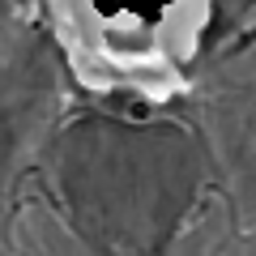}
  \caption{CSR}
\end{subfigure}
\begin{subfigure}{.13\textwidth}
  \centering
  \includegraphics[width=1\linewidth]{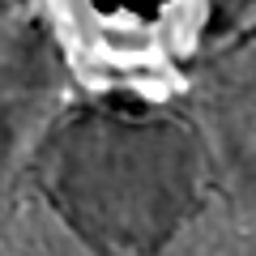}
  \caption{ULAP}
\end{subfigure}
\begin{subfigure}{.13\textwidth}
  \centering
  \includegraphics[width=1\linewidth]{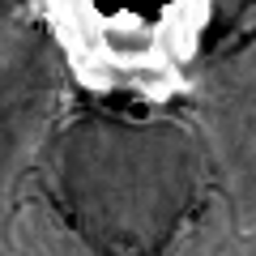}
  \caption{LP-CNN}
\end{subfigure}
\begin{subfigure}{.13\textwidth}
  \centering
  \includegraphics[width=1\linewidth]{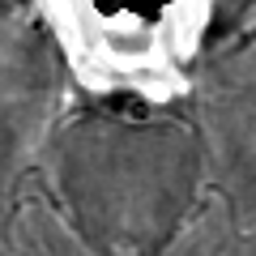}
  \caption{PCNN-NSST}
\end{subfigure}
\begin{subfigure}{.13\textwidth}
  \centering
  \includegraphics[width=1\linewidth]{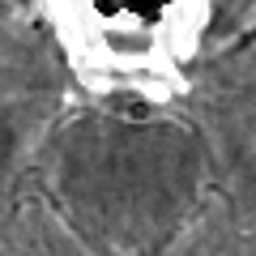}
  \caption{proposed}
\end{subfigure}
\caption{Example of MR(T2) -- CT fusion results.}
\label{fig: MRI-CT}
\end{figure*}

\begin{figure*}[h]
\centering
\begin{subfigure}{.13\textwidth}
  \centering
  \includegraphics[width=1\linewidth]{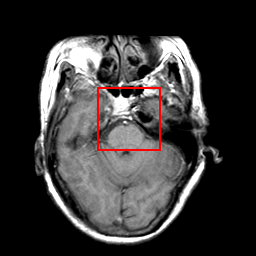}  
\end{subfigure}
\begin{subfigure}{.13\textwidth}
  \centering
  \includegraphics[width=1\linewidth]{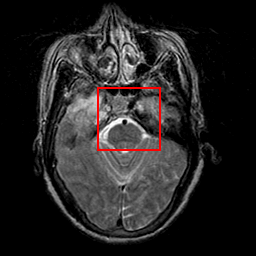}  
\end{subfigure}
\begin{subfigure}{.13\textwidth}
  \centering
  \includegraphics[width=1\linewidth]{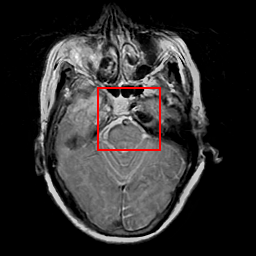}
\end{subfigure}
\begin{subfigure}{.13\textwidth}
  \centering
  \includegraphics[width=1\linewidth]{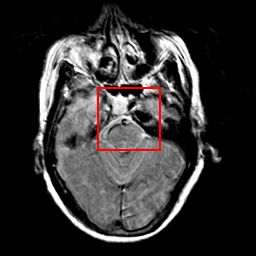}
\end{subfigure}
\begin{subfigure}{.13\textwidth}
  \centering
  \includegraphics[width=1\linewidth]{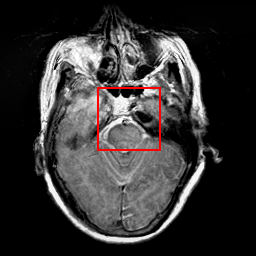}
\end{subfigure}
\begin{subfigure}{.13\textwidth}
  \centering
  \includegraphics[width=1\linewidth]{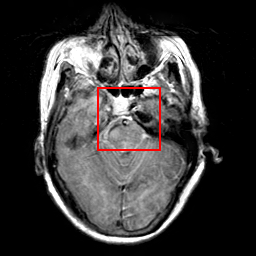}
\end{subfigure}
\begin{subfigure}{.13\textwidth}
  \centering
  \includegraphics[width=1\linewidth]{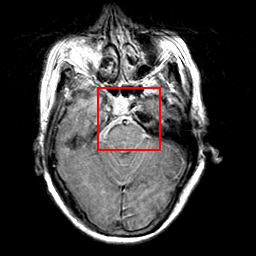}
\end{subfigure}

\vspace{1mm}

\begin{subfigure}{.13\textwidth}
  \centering
  \includegraphics[width=1\linewidth]{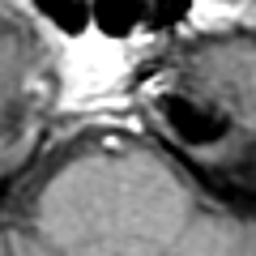}  
  \caption{MR(T1)}
\end{subfigure}
\begin{subfigure}{.13\textwidth}
  \centering
  \includegraphics[width=1\linewidth]{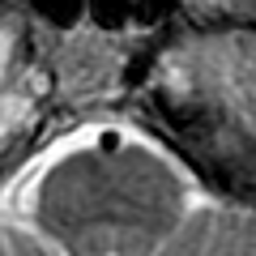}  
  \caption{MR(T2)}
\end{subfigure}
\begin{subfigure}{.13\textwidth}
  \centering
  \includegraphics[width=1\linewidth]{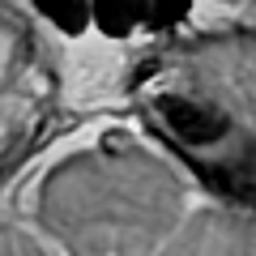}
  \caption{CSR}
\end{subfigure}
\begin{subfigure}{.13\textwidth}
  \centering
  \includegraphics[width=1\linewidth]{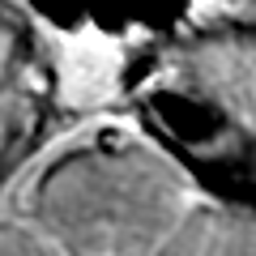}
  \caption{ULAP}
\end{subfigure}
\begin{subfigure}{.13\textwidth}
  \centering
  \includegraphics[width=1\linewidth]{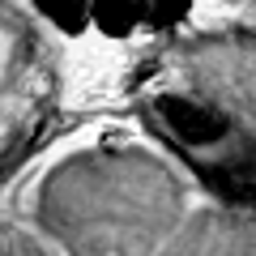}
  \caption{LP-CNN}
\end{subfigure}
\begin{subfigure}{.13\textwidth}
  \centering
  \includegraphics[width=1\linewidth]{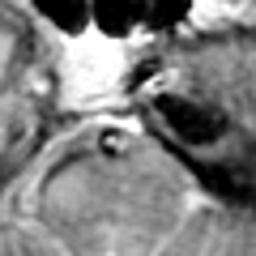}
  \caption{PCNN-NSST}
\end{subfigure}
\begin{subfigure}{.13\textwidth}
  \centering
  \includegraphics[width=1\linewidth]{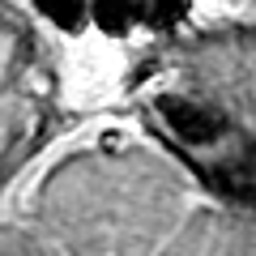}
  \caption{proposed}
\end{subfigure}
\caption{Example of MR(T1) -- MR(T2) fusion results.}
\label{fig: MRI-MRI}
\end{figure*}

\begin{figure*}[h]
\centering
\begin{subfigure}{.13\textwidth}
  \centering
  \includegraphics[width=1\linewidth]{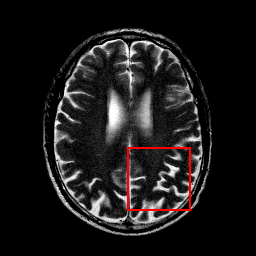}  
\end{subfigure}
\begin{subfigure}{.13\textwidth}
  \centering
  \includegraphics[width=1\linewidth]{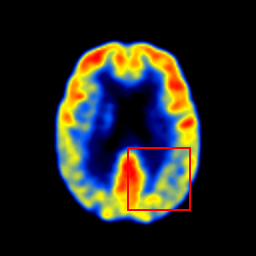}  
\end{subfigure}
\begin{subfigure}{.13\textwidth}
  \centering
  \includegraphics[width=1\linewidth]{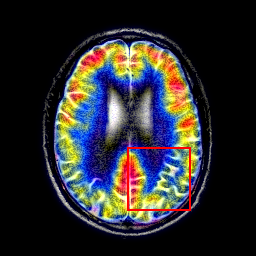}
\end{subfigure}
\begin{subfigure}{.13\textwidth}
  \centering
  \includegraphics[width=1\linewidth]{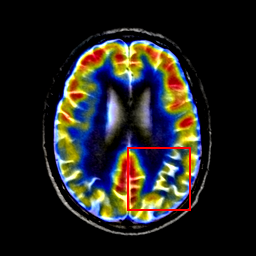}
\end{subfigure}
\begin{subfigure}{.13\textwidth}
  \centering
  \includegraphics[width=1\linewidth]{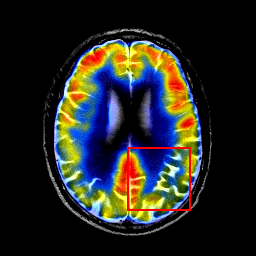}
\end{subfigure}
\begin{subfigure}{.13\textwidth}
  \centering
  \includegraphics[width=1\linewidth]{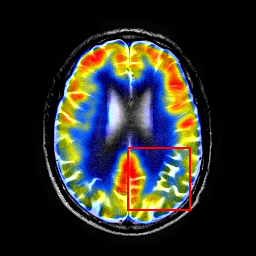}
\end{subfigure}
\begin{subfigure}{.13\textwidth}
  \centering
  \includegraphics[width=1\linewidth]{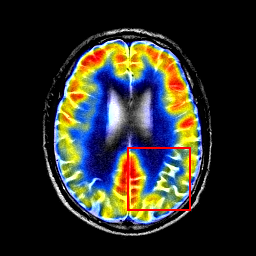}
\end{subfigure}

\vspace{1mm}

\begin{subfigure}{.13\textwidth}
  \centering
  \includegraphics[width=1\linewidth]{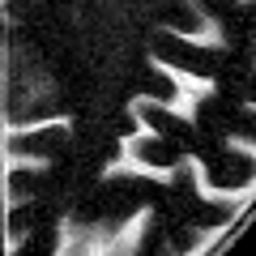}  
  \caption{MR(T2)}
\end{subfigure}
\begin{subfigure}{.13\textwidth}
  \centering
  \includegraphics[width=1\linewidth]{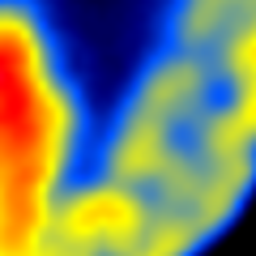}  
  \caption{PET}
\end{subfigure}
\begin{subfigure}{.13\textwidth}
  \centering
  \includegraphics[width=1\linewidth]{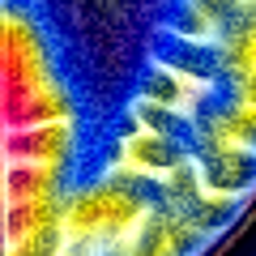}
  \caption{LLF-IOI}
\end{subfigure}
\begin{subfigure}{.13\textwidth}
  \centering
  \includegraphics[width=1\linewidth]{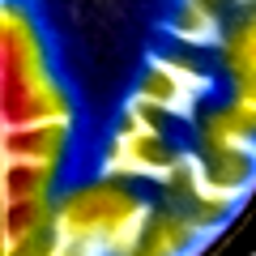}
  \caption{ULAP}
\end{subfigure}
\begin{subfigure}{.13\textwidth}
  \centering
  \includegraphics[width=1\linewidth]{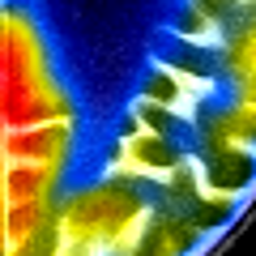}
  \caption{LP-CNN}
\end{subfigure}
\begin{subfigure}{.13\textwidth}
  \centering
  \includegraphics[width=1\linewidth]{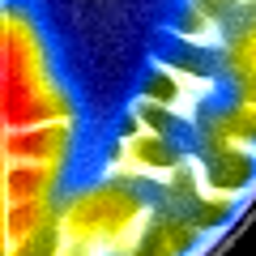}
  \caption{PCNN-NSST}
\end{subfigure}
\begin{subfigure}{.13\textwidth}
  \centering
  \includegraphics[width=1\linewidth]{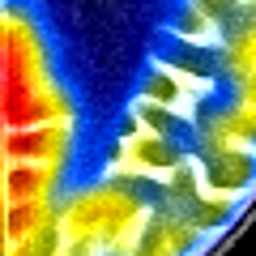}
  \caption{proposed}
\end{subfigure}
\caption{Example of MR(T2) -- PET fusion results.}
\label{fig: MRI-PET}
\end{figure*}

\begin{figure*}[h]
\centering
\begin{subfigure}{.13\textwidth}
  \centering
  \includegraphics[width=1\linewidth]{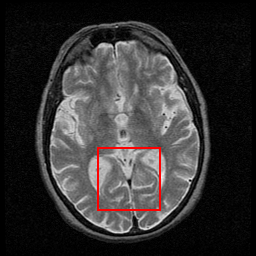}  
\end{subfigure}
\begin{subfigure}{.13\textwidth}
  \centering
  \includegraphics[width=1\linewidth]{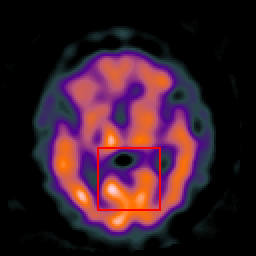}  
\end{subfigure}
\begin{subfigure}{.13\textwidth}
  \centering
  \includegraphics[width=1\linewidth]{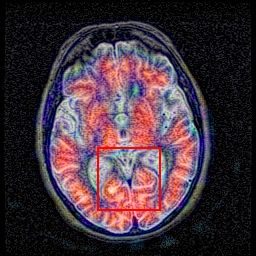}
\end{subfigure}
\begin{subfigure}{.13\textwidth}
  \centering
  \includegraphics[width=1\linewidth]{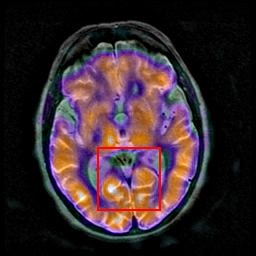}
\end{subfigure}
\begin{subfigure}{.13\textwidth}
  \centering
  \includegraphics[width=1\linewidth]{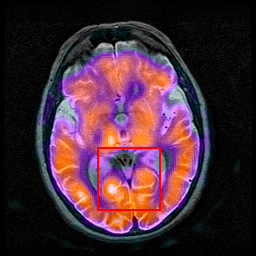}
\end{subfigure}
\begin{subfigure}{.13\textwidth}
  \centering
  \includegraphics[width=1\linewidth]{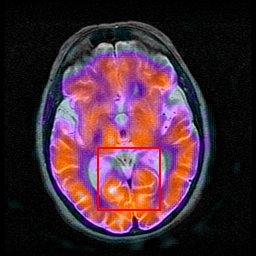}
\end{subfigure}
\begin{subfigure}{.13\textwidth}
  \centering
  \includegraphics[width=1\linewidth]{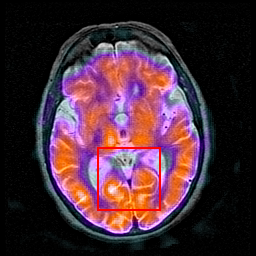}
\end{subfigure}

\vspace{1mm}

\begin{subfigure}{.13\textwidth}
  \centering
  \includegraphics[width=1\linewidth]{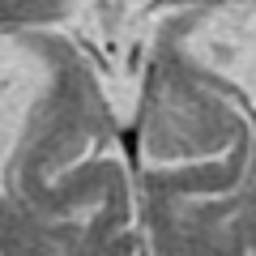}  
  \caption{MR(T2)}
\end{subfigure}
\begin{subfigure}{.13\textwidth}
  \centering
  \includegraphics[width=1\linewidth]{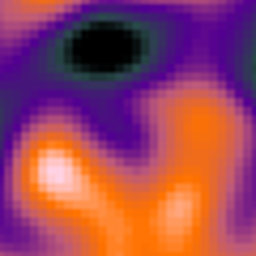}  
  \caption{SPECT(TC)}
\end{subfigure}
\begin{subfigure}{.13\textwidth}
  \centering
  \includegraphics[width=1\linewidth]{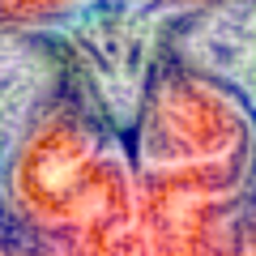}
  \caption{LLF-IOI}
\end{subfigure}
\begin{subfigure}{.13\textwidth}
  \centering
  \includegraphics[width=1\linewidth]{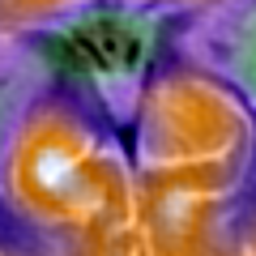}
  \caption{ULAP}
\end{subfigure}
\begin{subfigure}{.13\textwidth}
  \centering
  \includegraphics[width=1\linewidth]{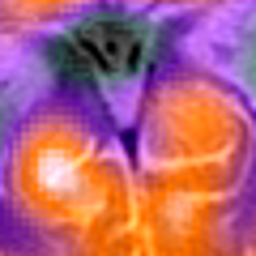}
  \caption{LP-CNN}
\end{subfigure}
\begin{subfigure}{.13\textwidth}
  \centering
  \includegraphics[width=1\linewidth]{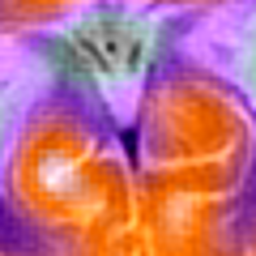}
  \caption{PCNN-NSST}
\end{subfigure}
\begin{subfigure}{.13\textwidth}
  \centering
  \includegraphics[width=1\linewidth]{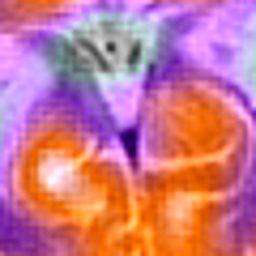}
  \caption{proposed}
\end{subfigure}
\caption{Example of MR(T2) -- SPECT(TC) fusion results.}
\label{fig: MRI-SPECT_TC}
\end{figure*}

\begin{figure*}[h]
\centering
\begin{subfigure}{.13\textwidth}
  \centering
  \includegraphics[width=1\linewidth]{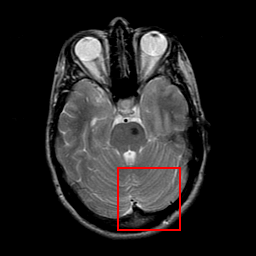}  
\end{subfigure}
\begin{subfigure}{.13\textwidth}
  \centering
  \includegraphics[width=1\linewidth]{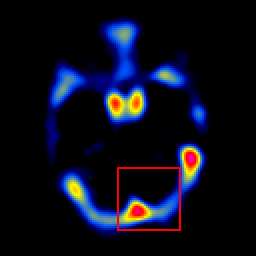}  
\end{subfigure}
\begin{subfigure}{.13\textwidth}
  \centering
  \includegraphics[width=1\linewidth]{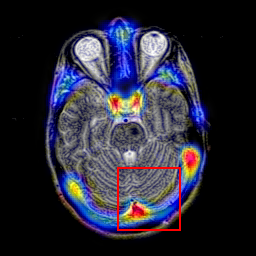}
\end{subfigure}
\begin{subfigure}{.13\textwidth}
  \centering
  \includegraphics[width=1\linewidth]{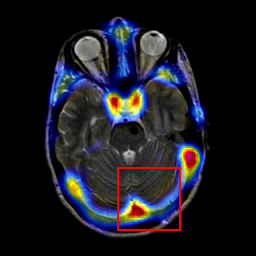}
\end{subfigure}
\begin{subfigure}{.13\textwidth}
  \centering
  \includegraphics[width=1\linewidth]{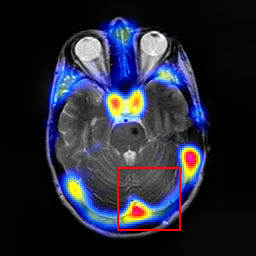}
\end{subfigure}
\begin{subfigure}{.13\textwidth}
  \centering
  \includegraphics[width=1\linewidth]{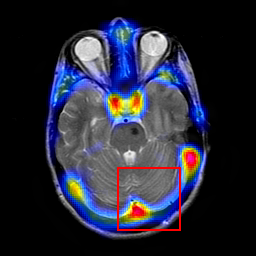}
\end{subfigure}
\begin{subfigure}{.13\textwidth}
  \centering
  \includegraphics[width=1\linewidth]{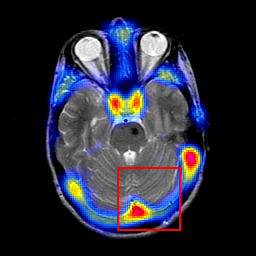}
\end{subfigure}

\vspace{1mm}

\begin{subfigure}{.13\textwidth}
  \centering
  \includegraphics[width=1\linewidth]{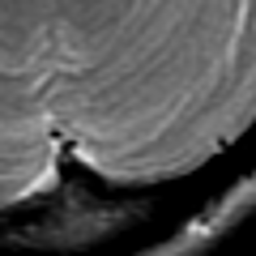}  
  \caption{MR(T2)}
\end{subfigure}
\begin{subfigure}{.13\textwidth}
  \centering
  \includegraphics[width=1\linewidth]{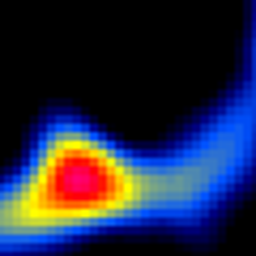}  
  \caption{SPECT(TI)}
\end{subfigure}
\begin{subfigure}{.13\textwidth}
  \centering
  \includegraphics[width=1\linewidth]{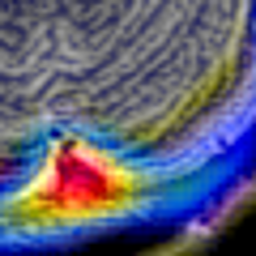}
  \caption{LLF-IOI}
\end{subfigure}
\begin{subfigure}{.13\textwidth}
  \centering
  \includegraphics[width=1\linewidth]{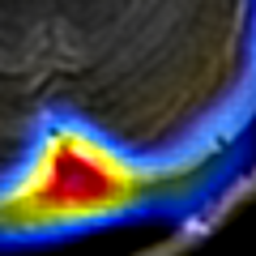}
  \caption{ULAP}
\end{subfigure}
\begin{subfigure}{.13\textwidth}
  \centering
  \includegraphics[width=1\linewidth]{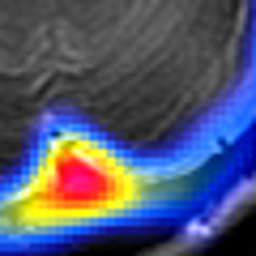}
  \caption{LP-CNN}
\end{subfigure}
\begin{subfigure}{.13\textwidth}
  \centering
  \includegraphics[width=1\linewidth]{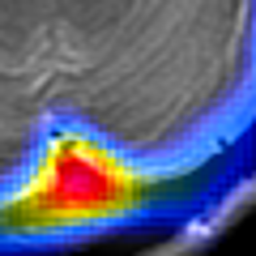}
  \caption{PCNN-NSST}
\end{subfigure}
\begin{subfigure}{.13\textwidth}
  \centering
  \includegraphics[width=1\linewidth]{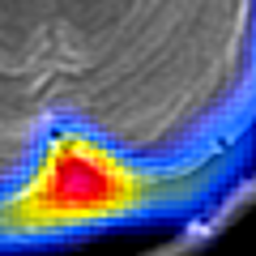}
  \caption{proposed}
\end{subfigure}
\caption{Example of MR(T2) -- SPECT(TI) fusion results.}
\label{fig: MRI-SPECTTI}
\end{figure*}

\begin{figure*}[h]
\centering
\begin{subfigure}{.13\textwidth}
  \centering
  \includegraphics[width=1\linewidth]{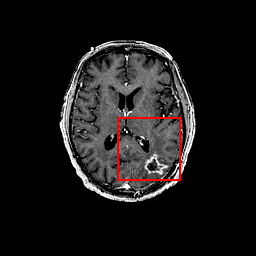}  
\end{subfigure}
\begin{subfigure}{.13\textwidth}
  \centering
  \includegraphics[width=1\linewidth]{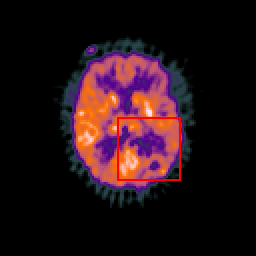}  
\end{subfigure}
\begin{subfigure}{.13\textwidth}
  \centering
  \includegraphics[width=1\linewidth]{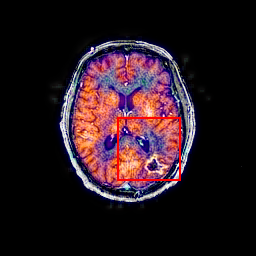}
\end{subfigure}
\begin{subfigure}{.13\textwidth}
  \centering
  \includegraphics[width=1\linewidth]{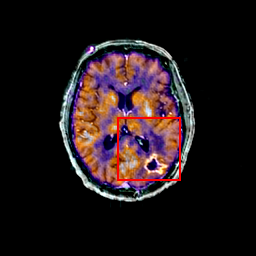}
\end{subfigure}
\begin{subfigure}{.13\textwidth}
  \centering
  \includegraphics[width=1\linewidth]{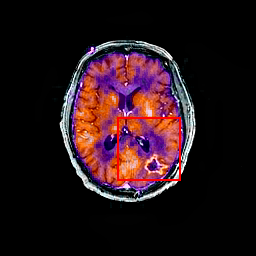}
\end{subfigure}
\begin{subfigure}{.13\textwidth}
  \centering
  \includegraphics[width=1\linewidth]{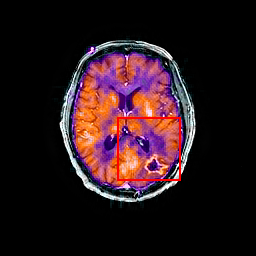}
\end{subfigure}
\begin{subfigure}{.13\textwidth}
  \centering
  \includegraphics[width=1\linewidth]{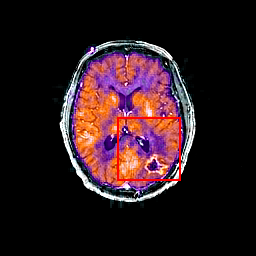}
\end{subfigure}

\vspace{1mm}

\begin{subfigure}{.13\textwidth}
  \centering
  \includegraphics[width=1\linewidth]{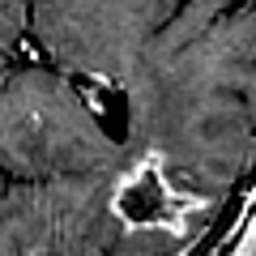}  
  \caption{MR(Gad)}
\end{subfigure}
\begin{subfigure}{.13\textwidth}
  \centering
  \includegraphics[width=1\linewidth]{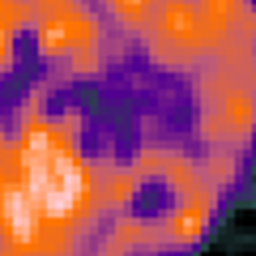}  
  \caption{PET}
\end{subfigure}
\begin{subfigure}{.13\textwidth}
  \centering
  \includegraphics[width=1\linewidth]{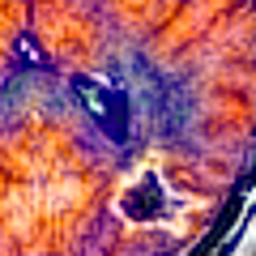}
  \caption{LLF-IOI}
\end{subfigure}
\begin{subfigure}{.13\textwidth}
  \centering
  \includegraphics[width=1\linewidth]{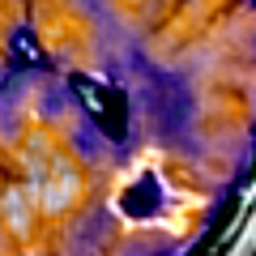}
  \caption{ULAP}
\end{subfigure}
\begin{subfigure}{.13\textwidth}
  \centering
  \includegraphics[width=1\linewidth]{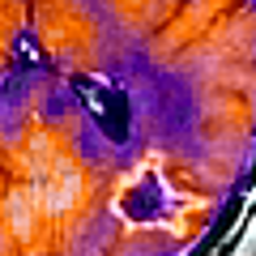}
  \caption{LP-CNN}
\end{subfigure}
\begin{subfigure}{.13\textwidth}
  \centering
  \includegraphics[width=1\linewidth]{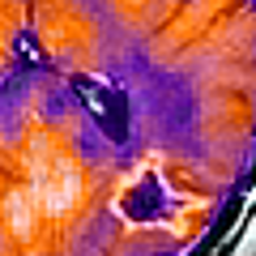}
  \caption{PCNN-NSST}
\end{subfigure}
\begin{subfigure}{.13\textwidth}
  \centering
  \includegraphics[width=1\linewidth]{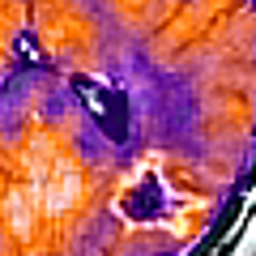}
  \caption{proposed}
\end{subfigure}
\caption{Example of MR(Gad) -- PET fusion results.}
\label{fig: GadMRI-Pet}
\end{figure*}

\subsection{Comparison with Other Methods}
\subsubsection{Visual Comparison}
Figs.~\ref{fig: MRI-CT}-\ref{fig: GadMRI-Pet} \NO{show the final fused image for one pair of images in each of the $6$ datasets.} \NO{In Figs.~\ref{fig: MRI-CT}, \ref{fig: MRI-MRI} and \ref{fig: MRI-SPECTTI}, one can see that} the ULAP and CSR methods \NO{clearly suffer from a loss of local intensity}. \NO{The same effect can be observed to a smaller extent for the LP-CNN method. This can be explained by the fact that these three methods rely on an} averaging-based approach for the fusion of low-resolution components, which usually contain most of the energy content of the images. \NO{For example, the intensities are significantly attenuated in regions where one of the input images is dark. The ULAP and LP-CNN methods also use averaging for their high-resolution components. In this case, averaging results in a loss of detail or texture, as can be seen in Figs.~\ref{fig: MRI-PET}(d,e) and \ref{fig: MRI-SPECTTI}(d,e).}  
%
\begin{table*}[ht]
\centering
\begin{tabular}{@{}ccccccccc@{}}
\toprule
Data-sets                           & Metrics               & CSR                    & LLF-IOI              & ULAP                   & LP-CNN                 & {\scriptsize PCNN-NSST} & proposed               \\ \midrule
\multirow{6}{*}{MR(T1)-MR(T2)}      & $Q_Y$                 & $0.8899$               & -                    & $0.8359$               & $0.7349$               & $0.8651$                & $\bb{0.9336}$          \\
                                    & $Q_{CB}$              & $0.7073$               & -                    & $0.6769$               & $0.5891$               & $0.7107$                & $\bb{0.7351}$          \\
                                    & $TMQI$                & $0.7745$               & -                    & $0.7680$               & $0.7694$               & $0.7768$                & $\bb{0.7813}$          \\
                                    & $STD$                 & $56.3661$              & -                    & $66.1464$              & $64.5726$              & $66.2631$               & $\bb{68.4629}$         \\ \midrule
\multirow{6}{*}{MR(T2)-CT}          & $Q_Y$                 & $0.8007$               & -                    & $0.7613$               & $0.7573$               & $0.7621$                & $\bb{0.8912}$          \\
                                    & $Q_{CB}$              & $0.6112$               & -                    & $0.5991$               & $0.5659$               & $0.5725$                & $\bb{0.6265}$          \\
                                    & $TMQI$                & $0.7285$               & -                    & $0.7228$               & $0.7101$               & $0.7277$                & $\bb{0.7447}$          \\
                                    & $STD$                 & $65.0380$              & -                    & $71.4757$              & $84.2350$              & $87.4317$               & $\bb{88.3130}$         \\ \midrule
\multirow{6}{*}{MR(T2)-PET}         & $Q_Y$                 & $0.6135$               & $0.8462$             & $0.8053$               & $0.8398$               & $0.8257$                & $\bb{0.8994}$          \\
                                    & $Q_{CB}$              & $0.6133$               & $0.6630$             & $0.6645$               & $0.6614$               & $0.6327$                & $\bb{0.6786}$          \\
                                    & $TMQI$                & $0.6985$               & $0.7391$             & $0.7298$               & $0.7341$               & $0.7382$                & $\bb{0.7438}$          \\
                                    & $STD$                 & $61.9247$              & $75.5607$            & $64.0611$              & $73.2243$              & $78.7324$               & $\bb{81.9925}$         \\ \midrule
\multirow{6}{*}{MR(T2)-SPECT(TC)}   & $Q_Y$                 & $0.5624$               & $0.7156$             & $0.7480$               & $0.7471$               & $0.8401$                & $\bb{0.8512}$          \\
                                    & $Q_{CB}$              & $0.5386$               & $0.6280$             & $0.6130$               & $0.5532$               & $0.6160$                & $\bb{0.6339}$          \\
                                    & $TMQI$                & $0.7023$               & $0.7359$             & $\bb{0.7409}$          & $0.7357$               & $0.7344$                &  $0.7384$              \\
                                    & $STD$                 & $60.6317$              & $65.2941$            & $57.3897$              & $67.4967$              & $67.8653$               & $\bb{70.4694}$         \\ \midrule  
\multirow{6}{*}{MR(T2)-SPECT(TI)}   & $Q_Y$                 & $0.6610$               & $0.7234$             & $0.7629$               & $0.7765$               & $0.8494$                & $\bb{0.8995}$          \\
                                    & $Q_{CB}$              & $0.6252$               & $0.4954$             & $0.5108$               & $0.5029$               & $0.5772$                & $\bb{0.5880}$          \\
                                    & $TMQI$                & $0.6778$               & $0.6956$             & $0.6958$               & $0.6932$               & $0.6941$                & $\bb{0.6959}$          \\
                                    & $STD$                 & $52.2014$              & $64.4325$            & $51.1641$              & $67.9101$              & $67.6377$               & $\bb{70.0668}$         \\ \midrule   
\multirow{6}{*}{MR(Gad)-PET}        & $Q_Y$                 & $0.6641$               & $0.7583$             & $0.8340$               & $0.7478$               & $0.8176$                & $\bb{0.9105}$          \\
                                    & $Q_{CB}$              & $0.6164$               & $0.6691$             & $0.6876$               & $0.6079$               & $0.6582$                & $\bb{0.7204}$          \\
                                    & $TMQI$                & $0.7232$               & $0.7468$             & $0.7461$               & $0.7396$               & $0.7463$                & $\bb{0.7508}$          \\
                                    & $STD$                 & $56.2280$              & $63.3500$            & $55.1129$              & $62.1655$              & $65.0489$               & $\bb{67.0991}$         \\ \bottomrule     
\end{tabular} 
\caption{Objective comparison between different methods. The best performance is shown in bold.} \label{table: objective results}
\end{table*}
\begin{table*}[h]
\centering
\begin{tabular}{@{}cccccccccc@{}}
\toprule
                        & CSR                  & LLF-IOI                & ULAP                 & LP-CNN                  & PCNN-NSST            & proposed             \\ \midrule
Average run-time (s)    & $34.57$                & $64.58$                  & $0.11$                 & $12.69$                & $4.62$                  & $6.43$                \\ \bottomrule     
\end{tabular}
\caption{Average execution times of different fusion methods.}
\label{table: runtime}
\end{table*}

\begin{figure*}[h]
\centering
\begin{subfigure}{1\textwidth}
  \centering
 \includegraphics[width=18cm]{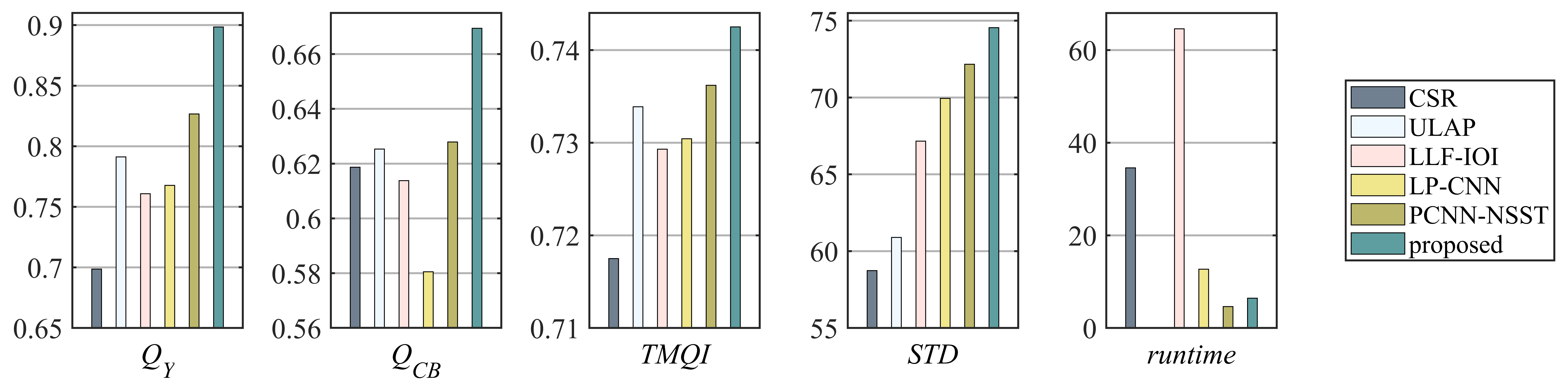} 
\end{subfigure}
\caption{Average objective evaluation results using all test datasets for different fusion methods.}
\label{fig: average objective}
\end{figure*}
\NO{All the results obtained using the LLF-IOI method contain clear artifacts and visual inconsistencies. This MST\hl{-based} method uses binary-selection-based fusion for both low and high-resolution features. As explained in Section~\ref{section_introduction}, binary selection of resolution-based components can cause loss of details and essential information. The same effect can be observed for the CSR method, which also relies on binary selection for its \textit{detail} layer corresponding to the highest resolution.} 

\NO{The PCNN-NSST method uses \hl{$49$} decomposition layers, which is significantly more than the CSR \hl{($2$ layers)} and LLF-IOI ($3$ layers) methods. Using a larger number of decomposition layers allows PCNN-NSST to capture more relevant information, including intensity and texture.} \hl{Also, the use of directional filters in NSST \NO{improves} the fusion of features with higher structural similarities.} \NO{However, the PCNN-NSST method occasionally suffers from a non-negligible amount of intensity attenuation, again due to employing a binary-selection rule for resolution-based features (see Figs.~\ref{fig: MRI-CT} and \ref{fig: MRI-MRI}.).} Moreover, \NO{the NSST reconstructs the final image solely based on a sparse representation, which inevitably leads to a loss of texture information. For example, the magnified region of Fig.~\ref{fig: MRI-SPECTTI} shows how the texture of the MR image appears with a significantly lower contrast in the PCNN-NSST result. Note that the corresponding regions in the SPECT(TI) image are entirely dark, meaning that the texture is expected to appear in the final image unaltered.}  
%
%
\begin{figure*}[h]
\centering
\begin{subfigure}{1\textwidth}
  \centering
 \includegraphics[width=18cm]{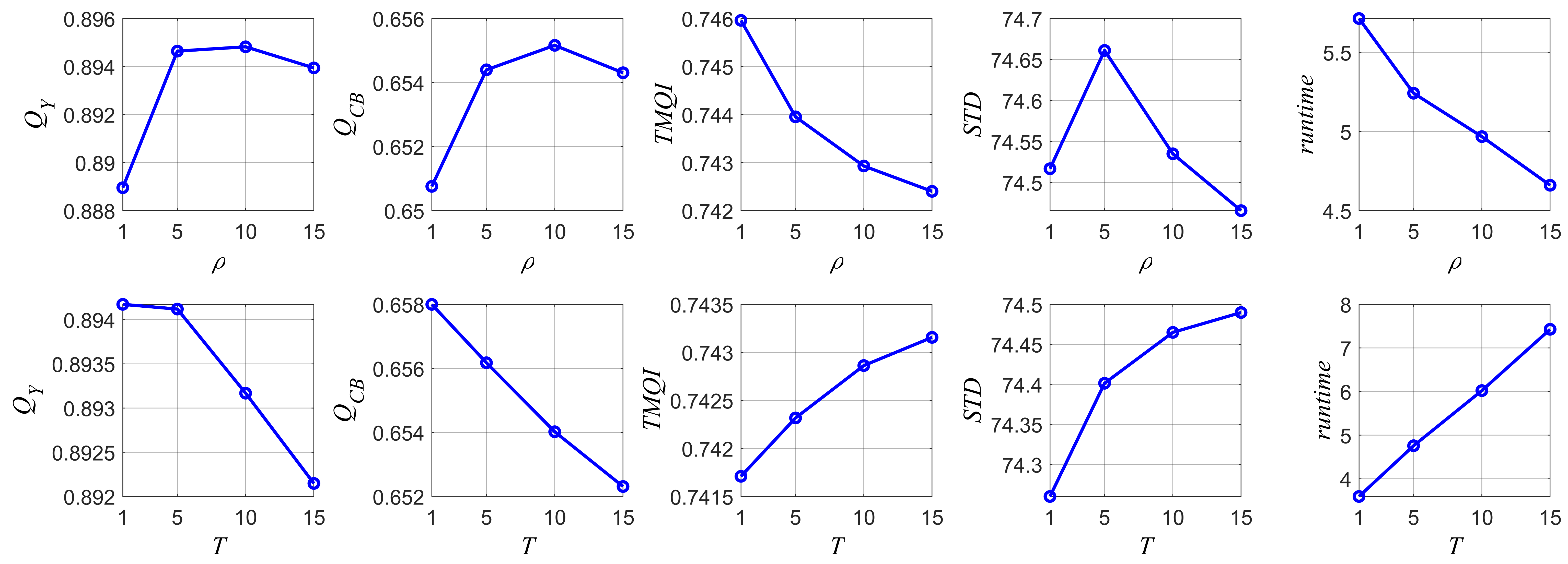} 
\end{subfigure}
\caption{Average objective evaluation results over all test datasets for different values of the penalty parameter $\rho$ and sparsity parameter $T$. Top: results for different $\rho$ values with $T=5$. Bottom: results for different $T$ values with $\rho=10$.}
\label{fig: param}
\end{figure*}
\NO{The proposed method provides good visual results by preserving both intensity and details. Recall that in the proposed method, the fusion rule is applied to the correlated features only, which guarantees that binary selection does not omit any modality-specific information.} Moreover, the independent components isolate these  modality-specific features and add them directly to the fused image without employing any \NO{additional} processing that might lead to \NO{texture degradation} or \NO{necessitate expensive} computations. The advantage of this strategy is evident in functional-anatomical fusion, \NO{where large areas in one of the images can be flat or dark in the other image.} For example, the regions \NO{of the MR image that are dark (no signal) in the PET image of Fig.~\ref{fig: MRI-SPECTTI} are well preserved by the proposed method, while all other methods show a loss of the local intensity or decreased contrast.} 
%
\subsubsection{Comparison Using Objective Metrics}
\NO{The results obtained based on the objective metrics are reported in} Table~\ref{table: objective results}. \NO{These results are in favour of the proposed method.} Specifically, \NO{the ULAP and CSR methods provide low STDs, which is due to the loss of contrast discussed previously.} \hl{Moreover, the LLF-IOI method always shows relatively low $Q_Y$ and $Q_{CB}$ values, which points to the presence of visual artefacts}. The objective metrics of the LP-CNN method are \hl{almost always} lower than those of the PCNN-NSST and proposed methods. Finally, the proposed method leads to the best overall performance \NO{(except one experiment where the performance is slightly below that of the ULAP method).} \NO{These results are summarized in terms of average objective metrics in Fig.~\ref{fig: average objective}, where one can see that the proposed method leads to the best average values over all datasets. These findings show that the proposed method generalizes well to diverse medical imaging modalities.} 
\subsubsection{Execution Times} \label{section_executiontimes}
\NO{The average execution times of all the experiments are reported in} Table~\ref{table: runtime}. This table shows that the proposed method is competitive with recent multimodal fusion methods in terms of computational efficiency. Specifically, the running time of the proposed method is comparable to that of the PCNN-NSST method and significantly better than those of the CSR \hl{and} LLF-IOI methods. The ULAP method results in the shortest execution time but does not yield the best results.

\subsection{Effect of Parameters}\label{section_parameffect}
\NO{Because the tested medical imaging modalities are very different, the parameters providing the best performance may vary between experiments. For example, fusion problems involving one low-detail modality, such as MR-SPECT, are efficiently tackled by a relatively small sparsity parameter $T$.  In contrast, the MR-CT fusion problem naturally requires higher values for $T$ to accurately represent the richer details in both modalities. However, in this work, the parameters used for the proposed method are not tuned for each pair of modalities separately. Instead, we use the parameters providing the best average performance over all datasets. This choice allows us also to test the generalisation ability of the parameter setting of the proposed method.}
\subsubsection{Dictionary Parameters}
\NO{The best patch size is related to the size of local features in the input images. The patch size also impacts the computational cost of the fusion problem. We use a value of $m=64$ in the presented experiments in order to achieve a good compromise between running time and \NO{effective capturing of features}. The number of dictionary atoms is set to $n=128$, although using smaller values of $n$ in the anatomical-functional image fusion problems (\textit{e.g.}, $n=64$) are sufficient to achieve the same performance (with significantly lower computational cost). As mentioned above, we choose the parameters that provide the best performance across all modalities.} 
\subsubsection{Sparsity Level and Penalty Parameter} 
\NO{The relationship between the penalty $\rho$ and sparsity level $T$ is explored in Fig.~\ref{fig: param}. One can see that increasing $\rho$ and increasing $T$ have the opposite effect on the evolution of the objective metrics. A large $\rho$ reduces the influence of the Pearson correlation term in the proposed cost function, which allows $\boldsymbol{E}_1$ and $\boldsymbol{E}_2$ \ff{to absorb large parts of $\boldsymbol{X}_1$ and $\boldsymbol{X}_2$} (\textit{i.e.,} they become increasingly sparse). Conversely, a small $T$ enforces higher sparsity of \ff{$\boldsymbol{Z}_1$ and $\boldsymbol{Z}_2$}, leading to more features being captured in the independent components $\boldsymbol{E}_1$ and $\boldsymbol{E}_2$. Note that the execution times also present a reverse behaviour when increasing $\rho$ or $T$.}

\NO{Fig.~\ref{fig: param} also highlights some challenges regarding the parameter setting. Specifically, the objective metrics $Q_Y$ and $Q_{CB}$ evolve in a manner that is inversely proportional to $TMQI$ and $STD$. More specifically, the former become worse when increasing $T$/lowering $\rho$, while the latter are improved.  Although decreasing the amount of information captured by the independent components does lead to loss of details; it also increases the amount of energy left in the correlated components. Following similar reasoning, lowering $T$/increasing $\rho$ saturates the independent components and leads to a loss of contrast. (Note that when the values in \ff{the final fused image} exceed $1$, they are simply rescaled by clipping them to the valid intensity range.) This trade-off should be taken into account when setting the parameters of the proposed method.}

Based on the observations above, \NO{the most suitable parameters are $T=5$ and $\rho=10$ in the presented experiments.} 
\section{Conclusion}
A novel image \NO{fusion} method for multimodal medical images. A decomposition method \NO{separates} input images into their correlated and independent components. The correlated components are captured by \NO{sparse representations with identical supports and learnt coupled dictionaries}. The independence between the independent components is established by minimization of pixel-wise Pearson correlations. An alternating optimization strategy is adopted for addressing the resulting optimization problem. \NO{One particularity of the proposed method is that it applies a fusion rule to the correlated components only while fully preserving the independent components. In the experiments, this strategy has shown superior preservation of intensity and detail compared to other recent methods. Quantitative evaluation metrics and comparison of execution times have also shown the competitiveness of the proposed method.}
\label{section_conclusion}
\bibliographystyle{IEEEtran}
\bibliography{MMIF}
\end{document}